\documentclass[10pt,twocolumn,letterpaper]{article}

\usepackage{cvpr}
\usepackage{times}
\usepackage{epsfig}
\usepackage{graphicx}
\usepackage{amsmath}
\usepackage{amssymb}
\usepackage{subcaption}
\usepackage{color}
\usepackage{authblk}

\usepackage[breaklinks=true,bookmarks=false]{hyperref}

\cvprfinalcopy 

\def\cvprPaperID{****} 

\begin{document}

\title{Single Image Deraining using Scale-Aware Multi-Stage Recurrent Network}

\author[1]{Ruoteng Li}
\author[1]{Loong-Fah Cheong}
\author[1,2]{Robby T. Tan}
\affil[1]{National University of Singapore} \affil[2]{Yale-NUS}

\maketitle

\addtocounter{footnote}{1}

\begin{abstract}
Given a single input rainy image, our goal is to visually remove rain streaks and the veiling effect caused by scattering and transmission of rain streaks and rain droplets. We are particularly concerned with heavy rain, where rain streaks of various sizes and directions can overlap each other and the veiling effect reduces contrast severely. To achieve our goal, we introduce a scale-aware multi-stage convolutional neural network. Our main idea here is that different sizes of rain-streaks visually degrade the scene in different ways. Large nearby streaks obstruct larger regions and are likely to reflect specular highlights more prominently than smaller distant streaks. These different effects of different streaks have their own characteristics in their image features, and thus need to be treated differently. To realize this, we create parallel sub-networks that are trained and made aware of these different scales of rain streaks. To our knowledge, this idea of parallel sub-networks that treats the same class of objects according to their unique sub-classes is novel, particularly in the context of rain removal. To verify our idea, we conducted experiments on both synthetic and real images, and found that our method is effective and outperforms the state-of-the-art methods.
\end{abstract}

\section{Introduction}
Rain, particularly heavy rain, can impair visibility considerably. Individual rain streaks mar object's appearance with their specular highlights, refraction, scattering and blurring effects. Distant rain streaks accumulated along the line of sight degrade the visibility of a background scene by creating fog-like veiling effect. Moreover, as a result of the projection process, rain streaks in the image have different sizes and densities, depending on their distances to the camera; these different rain layers with different sizes and densities are confusedly overlaid upon each other in the image, severely aggravating the problem. All these can affect the performance of current computer vision algorithms, particularly those assuming clear visibility.

\begin{figure}
    \includegraphics[width=0.495\linewidth]{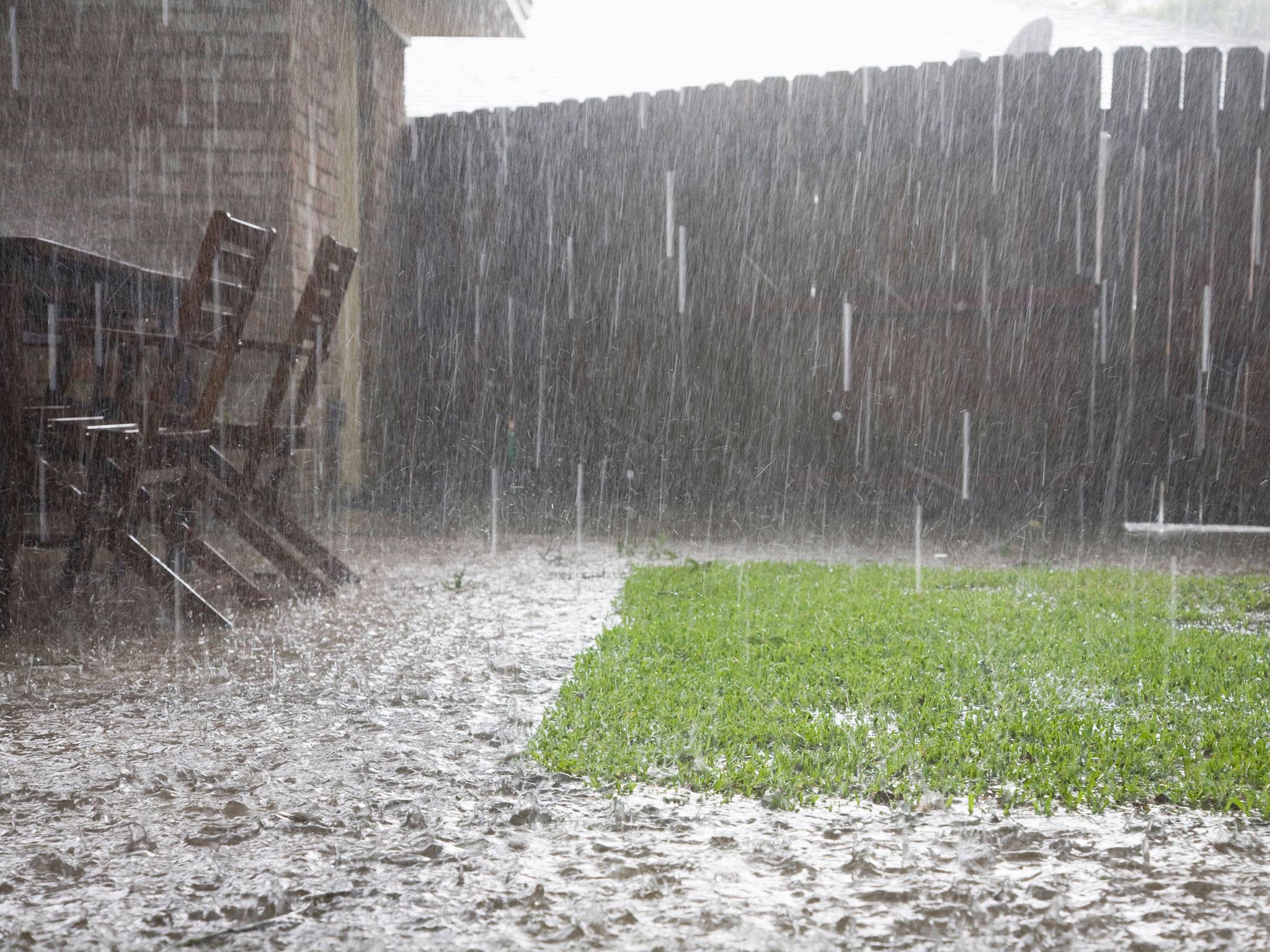}
    \includegraphics[width=0.495\linewidth]{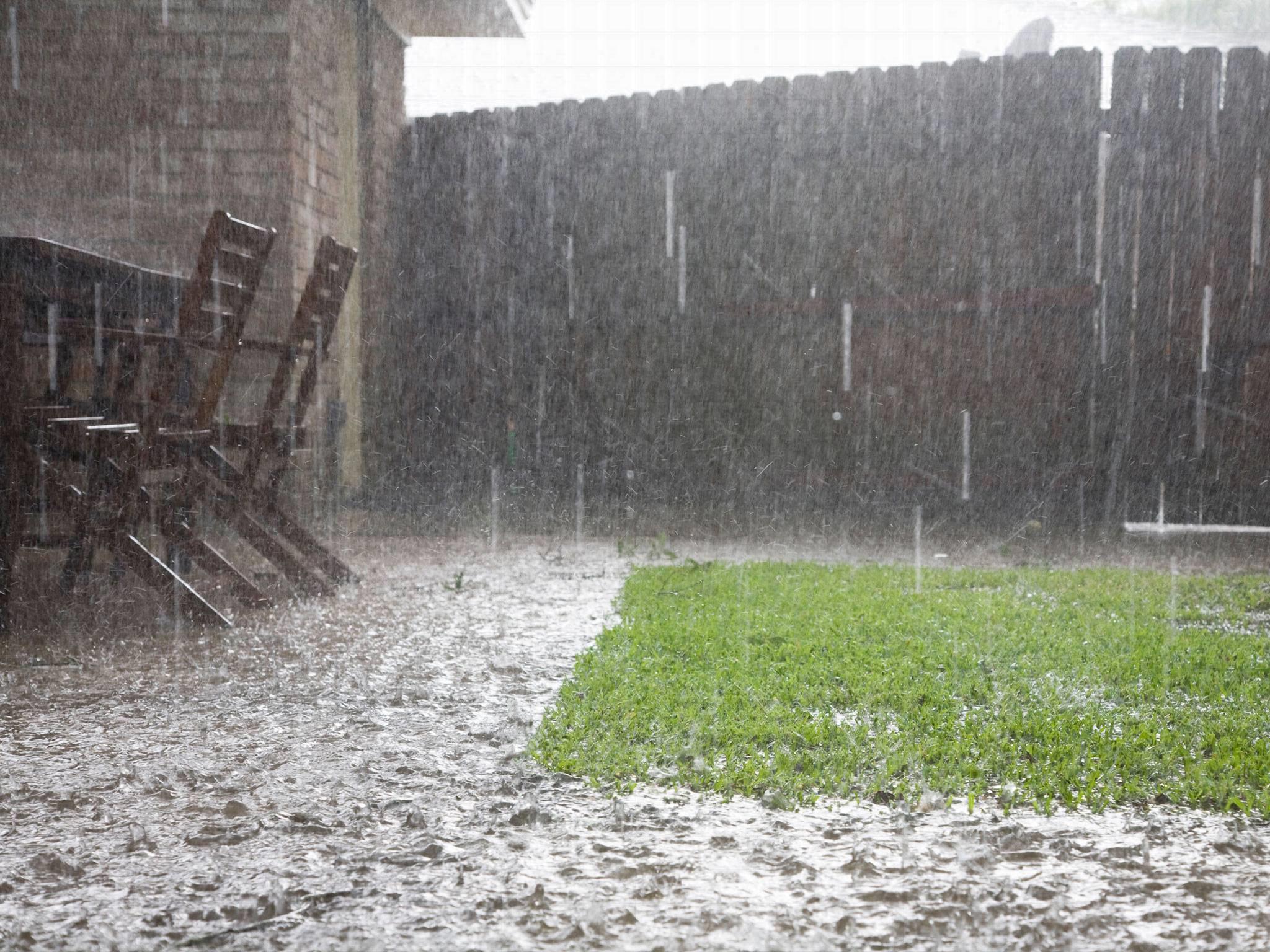}
    \includegraphics[width=0.495\linewidth]{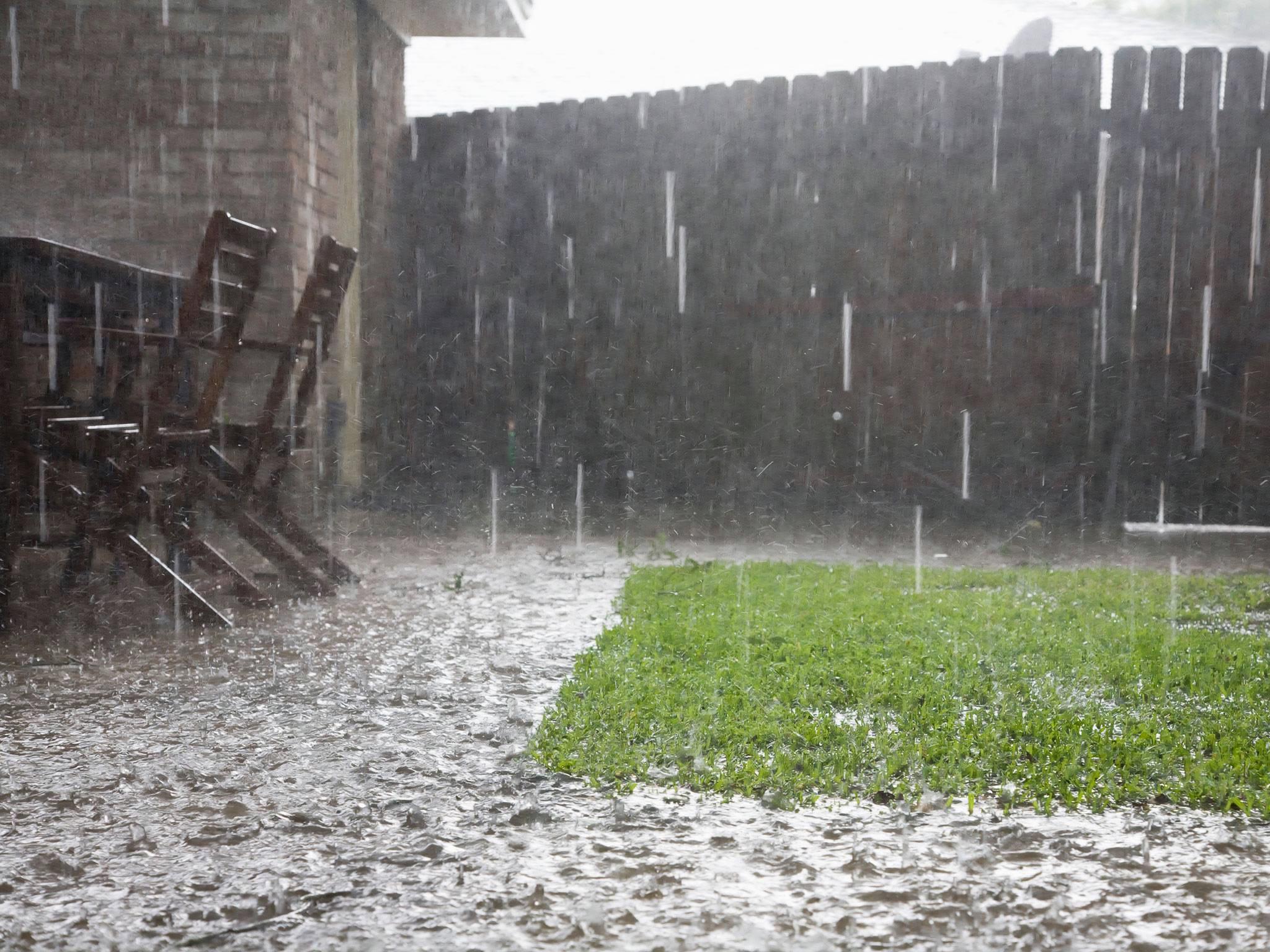}
    \includegraphics[width=0.495\linewidth]{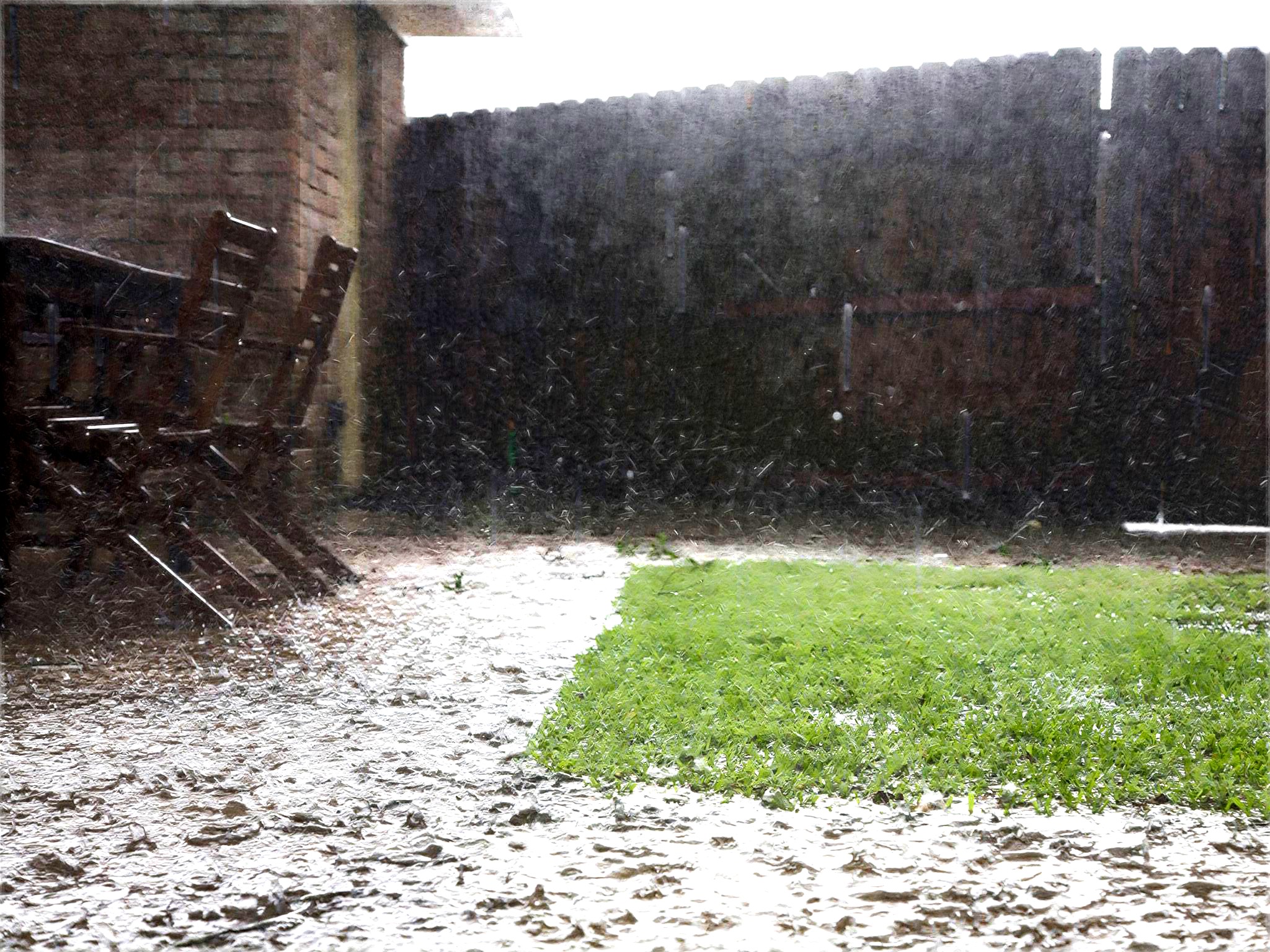}
    \caption{ Comparison of our method with state-of-the-art deraining algorithms. \textbf{Top Left}: Input image. \textbf{Top Right}: Result of \cite{Yang_2017_CVPR}. \textbf{Bottom Left}: Result of \cite{Fu_2017_CVPR}. \textbf{Bottom Right}: Result of the proposed method. }
    \label{fig:Cover}
\end{figure}

A few  rain streaks removal techniques have been proposed in the past decade to eliminate the rain streaks and restore visibility. Video-based rain streak removal methods (e.g.   \cite{Garg:2006}\cite{Kim_2015_TIP}) focus on image recovery from video sequence by exploring temporal information and frequency properties of rain. Some single-image based rain removal methods regard the problem as blind signal separation problem (e.g. \cite{Li_2016_CVPR}). Recently, deep learning methods have been applied to this area and have demonstrated their advantages in recovering background scenes from rainy images \cite{Yang_2017_CVPR}\cite{Fu_2017_CVPR}\cite{DBLP:journals/corr/FuHDLP16}. While all the aforementioned methods have demonstrated some degree of success, it is fair to say that they have not been subject to the full force of the tropical heavy rain and been tested where the scenes contain a range of depths. Both these factors render the deraining problem much harder; not only do we need to deal with a diverse range of rainfall from slight drizzle to an almost solid curtain of water, the rain might appear thinner in the foreground but thicker in the more distant areas in the same image. To make deraining algorithms resilient to these more severe conditions, there are several limitations of current algorithms that need to be carefully addressed.

\begin{figure}
    \includegraphics[width=1.0\linewidth]{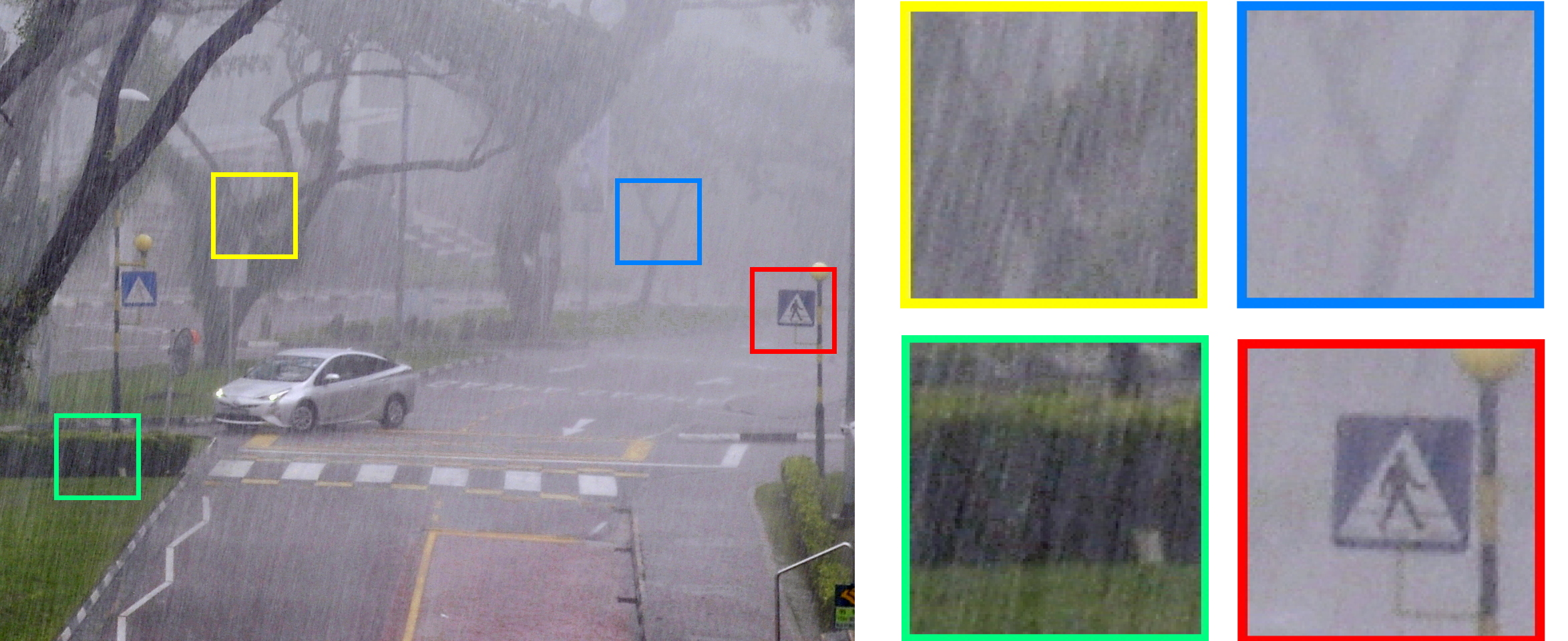}
    \caption{An example of heavy rain image. \textbf{Green} window: The compounding result of rain streaks accumulation and veiling effect. \textbf{Yellow} window: Rain streaks of various size overlap on each other. \textbf{Blue} and \textbf{Red} window: The fog-like veiling effect varies according to the object depth. Further object has stronger veiling effect. }
    \label{fig:RainExample}
\end{figure}

\begin{figure}
    \includegraphics[width=0.325\linewidth]{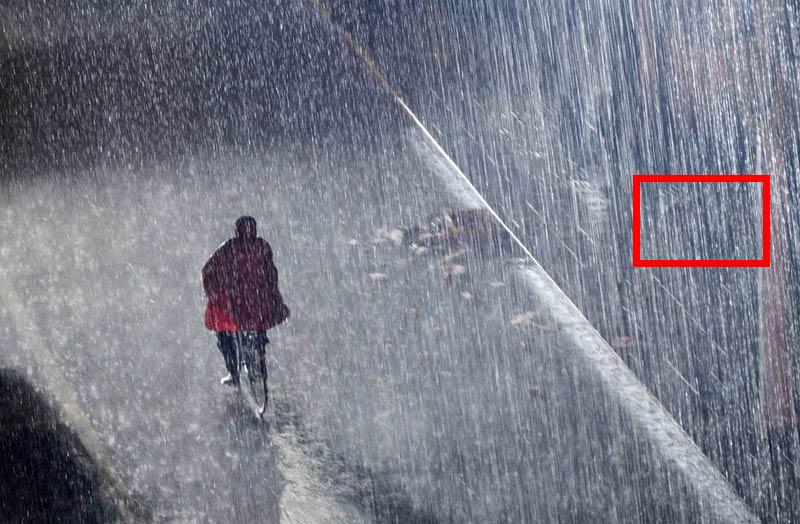}
    \includegraphics[width=0.325\linewidth]{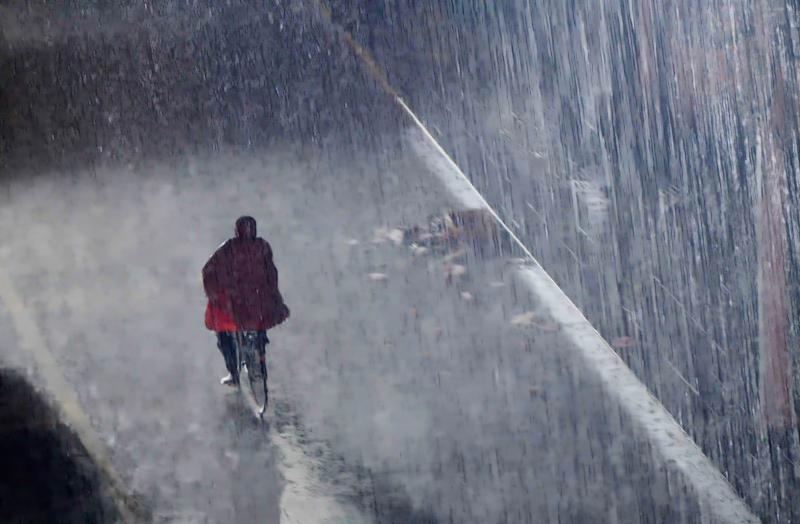}
    \includegraphics[width=0.325\linewidth]{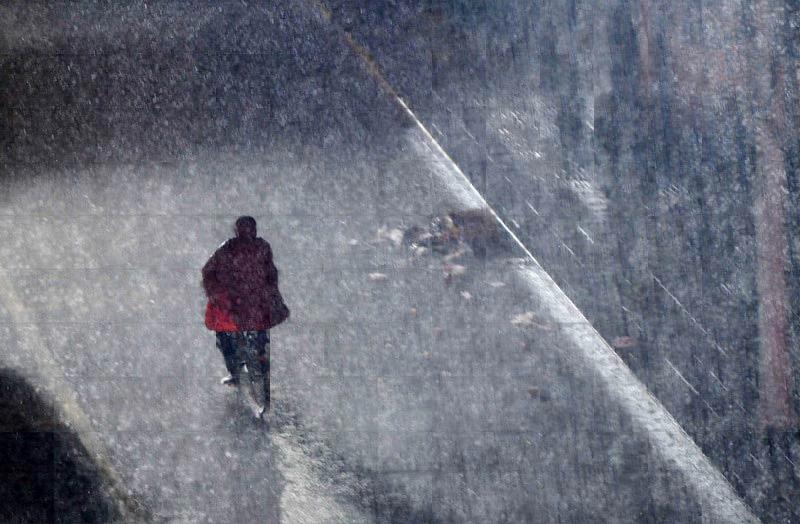}

    \includegraphics[width=0.325\linewidth]{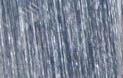}
    \includegraphics[width=0.325\linewidth]{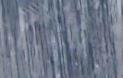}
    \includegraphics[width=0.325\linewidth]{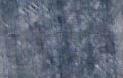}
    \caption{An example of heavy rain accumulation and results of two state-of-the-art rain removal methods. \textbf{Left}: Input rainy image. \textbf{Middle}: Results of \cite{Li_2016_CVPR}. Though this method removes some small-sized rain streaks, it cannot remove most of the long streaks effectively. \textbf{Right}: Results of \cite{Yang_2017_CVPR}. This method removes thinner rain streaks, but the thicker and wider rain streaks are left behind. }
    \label{fig:Overlap}
\end{figure}

First of all, most of the existing rain streaks removal methods (\cite{Luo_SparseCoding}\cite{Kang12Rain}\cite{Li_2016_CVPR}) are developed based on an assumption that the rain streaks distributed on a captured image is sparse. However, in the real world, particularly in the case of heavy rain or even in the case of moderate rain, if the scene extends far enough in depth, the dense rain streaks accumulation makes this assumption invalid as shown in Fig.~\ref{fig:RainExample}. Second, rain streaks of different sizes overlapping  each other can cause ambiguity or even unintelligibility for feature-based and learning-based methods as shown in Fig.~\ref{fig:Overlap} \cite{Yang_2017_CVPR}\cite{Li_2016_CVPR}. Although the appearance of an individual rain streak follows a rain model as described in \cite{Garg:2006}, the large number of streaks overlapping in various sizes and densities at best significantly expands the feature space to learn, and at worst produces a rain image comprising of several scales of phenomena interpenetrating one another, rendering the existing feature-based learning methods \cite{Kang12Rain}\cite{chen2013generalized}\cite{Luo_SparseCoding}\cite{Li_2016_CVPR}\cite{Yang_2017_CVPR} inefficient in correctly detecting rain streaks.
Finally,  the atmospheric veiling effect caused by light scattering process of both tiny and large rain droplets plays an important role in degrading the visibility of a rainy scene (Fig.\ref{fig:RainExample}). However most deraining algorithms did not address this problem properly.

Considering the aforementioned limitations, our goal is to develop a novel method that is capable of removing rain streaks and rain accumulation from a single image under various conditions, ranging from slight rain to heavy rain, and thus enhancing the visibility of the image. To achieve our goal, we introduce a multi-stage scale-aware convolutional neural network.
Our main idea here is that different sizes of rain-streaks visually degrade the scene in different ways. Large nearby streaks obstruct larger regions and are likely to reflect specular highlights more prominently than smaller distant streaks  \cite{Tamburo2014}. These different effects of different streaks have their own characteristics in their image features, and thus need to be treated differently. Thick rain streaks also tend to be lower in density and thus need larger spatio-temporal windows to properly analyze them. One might think that the different layers of a deep learning algorithm might be able to do this automatically, but when the differently-sized rain streaks are so inextricably mixed together, we contend that this is far from being the cases. To realize the different treatments effectively, we create parallel sub-networks that are trained and made aware of these different scales of rain streaks. To our knowledge, this idea of parallel sub-networks that treat the same class of objects according to their unique sub-classes is novel, particularly in the context of rain removal.

Our scale-aware multi-stage convolutional neural network consist of three parts. First, we adopt  DenseNet \cite{Huang_2017_CVPR}  as a backbone to extract general features. Next to it is parallel  sub-networks, each of which is trained to estimate rain streaks intensity map at a scale. These parallel sub-networks are recurrent convolutional layers with shortcut connections \cite{He_2015_CVPR}, which are iteratively refined to produce better rain streaks predictions. Then the input image subtracts the summed results of all the subnetworks and feeds forward to the next stage of parallel recurrent sub-networks. In each of the subsequent stages, the recurrent subnetworks predict the residual rain streaks based on the proceeding subtraction results from previous stage. Finally, the estimated rain streak maps will be combined with the input image to restore a clean background scene.

\begin{figure}
    \includegraphics[width=1.0\linewidth]{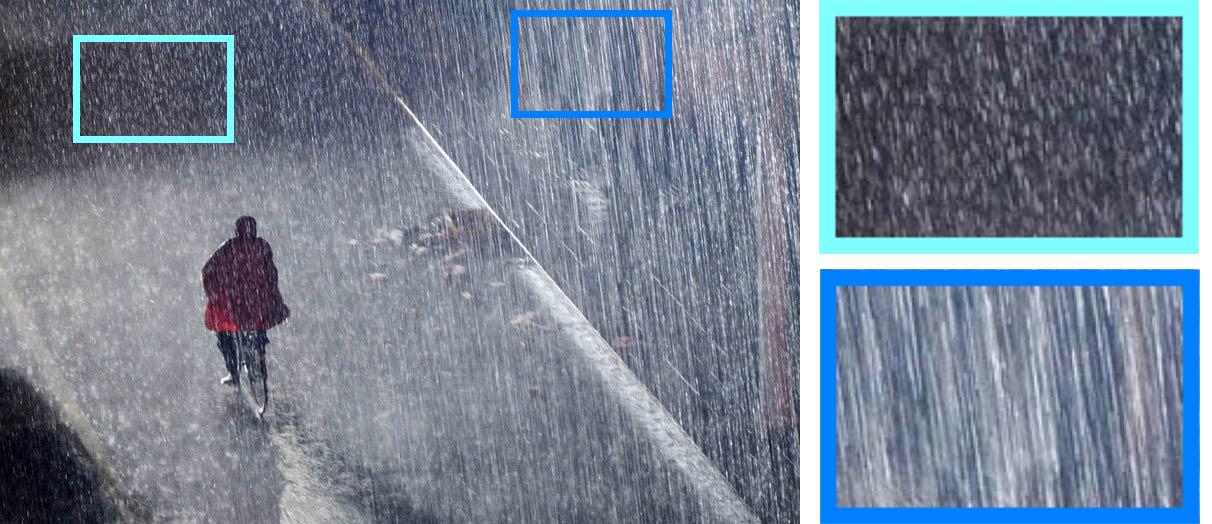}
    \caption{ An example of real rain image. Rain streaks of different sizes may appear at the same time, and overlap each other.  The enlarged windows demonstrate the different features of the rain streaks. }
    \label{fig:RainStreakComparison}
\end{figure}

Our key contributions can be summarized as follows:

\begin{enumerate}
\item We introduce an end-to-end network to remove both the rain streaks and their accumulation effect.
\item The proposed scale-aware network addresses the overlapping rain streaks of different sizes and densities using parallel recurrent sub-networks. Each sub-network is trained to extract rain streak features at a certain range of rain-streak sizes, decomposing the task of learning various streak sizes into learning smaller ranges of sizes. The reason of doing this is twofold. Firstly, different size of rain streaks is visually different and thus has different image features as shown in Fig.~\ref{fig:RainStreakComparison}. Secondly, these differently-sized rain streaks manifest themselves at different densities --- \eg thick rain streaks occur more sparsely --- and thus the scale of analysis for each subtask should be different.
\item Our method is able to remove the veiling effects created by atmospheric light scattering process of the accumulated rain droplets. A few CNN-based image enhancement methods (e.g. \cite{dehaze}\cite{Yang_2017_CVPR}) require post-processing and suffer from darkened output results. Based on our introduced rain model, our new formulation and network architecture allow our network to do end-to-end training and to recover the sharp background with brighter and richer preserved details.
\item Our network outperforms the state-of-the-art deraining methods on both synthetic and real rain datasets.
\end{enumerate}

In our investigation, we find that using DenseNet \cite{Huang_2017_CVPR} to extract general features improves the deraining quality significantly. Comparing the results with those of simple shallower convolutional networks, our experiments show the deeper network structure performs better.


\section{Related Works}
There are a number of methods proposed to improve the visibility of rain images, and we can categorize them into video-based
and single-image based methods.

\paragraph{Video Based Methods}
Early rain streaks removal methods focus on rain removal from video. Garg and Nayar's \cite{Garg:2006} assumes the background scene to be static and explores the temporal information of dynamic rain streaks to detect the streak location, where the intensity change is larger than a predefined threshold. Having detected the rain streaks, it further removes the rain streaks by taking the average intensity of the pixels taken from the previous and subsequent frames. This method is applicable only on static scenes. Bossu et al.'s \cite{Bossu2011} proposes a rain detection algorithm based on the concept of foreground-background separation. The foreground model is used to detect rain streaks by applying selection rules based on the photometric properties of rain, which assumes a raindrop is a moving object brighter than the background. A histogram of orientation of rain streaks (HOS) is used to reject those detected pixels that do not correspond to rain streaks.  Kim et al.'s \cite{Kim_2015_TIP} utilizes optical flow and only needs three successive frames to detect rain and also differentiate rain from other moving objects. It obtains the initial rain map by comparing a frame with the warped image of the subsequent frame using optical flow. The initial map is decomposed into valid rain streaks and non-rain streaks. Similar method utilizing temporal information has been applied to image restoration under water \cite{DBLP:conf/iccv/TianN09}.

\paragraph{Single-Image Based Methods}
For single-image rain streak removal, Kang et al.'s \cite{Kang12Rain} introduces a method that decomposes an input image into its low frequency component (structure layer) and a high-frequency component (texture layer). The high-frequency layer, which contains rain streaks, is used to extract rain streaks component and background details using sparse-coding based dictionary learning.  Luo et al's \cite{Luo_SparseCoding} proposes a discriminative sparse coding framework. Its objective function employs a dictionary, its background coefficients and rain layer coefficients. The goal of the objective function is to learn the background and rain layer by forcing the coefficient vector to be sparse.

Li et al's \cite{Li_2016_CVPR} decomposes the rain image into rain-free background layer and rain streak layer by utilizing Gaussian Mixture Models (GMMs) as a prior of background and rain streaks layers. The GMM prior for the background is learned from natural images, while that for the rain layer is learned from the input rain image. Fu et al's \cite{Fu_2017_CVPR} is a deep convolutional network that is based on Kang et al's idea \cite{Kang12Rain}. The network receives high-frequency component of an input rainy image and learns the negative residual map of rain streaks. The output of the network is then added back to the low-frequency component of the input rainy image to restore the clean background. Yang et al's \cite{Yang_2017_CVPR} is a CNN based method that learns to detect and remove the rain streaks simultaneously from a single image. The network uses a contextualized dilated network to learn a pool of features. From the features, a binary rain region mask is learnt to detect the rain streak location. Subsequently, the rain region mask is fed into the network to further learn rain streak intensity, and hence the clean image can be restored by subtracting the rain intensity map from input rain image.

\section{Rain Models}

The widely used rain model describes  the observed rain image \( \mathbf{O}\) as a linear combination of the rain-free background scene \( \mathbf{B}  \) and the rain streak layer \( \mathbf{R}  \) \cite{Luo_SparseCoding}\cite{Li_2016_CVPR}\cite{Huang_2012_ICM} :

\begin{equation}
    \mathbf{O} = \mathbf{B} + \mathbf{R}.
    \label{eq:LinearSuperposition}
\end{equation}
The objective for any rain streak removal algorithm is to remove the rain streaks layer \(\mathbf{R}\) from the rainy image \(\mathbf{O}\) to obtain the background scene \( \mathbf{B}\).

Rain removal algorithms based on Eq.~(\ref{eq:LinearSuperposition}) assume the rain streaks sparse and utilize individual rain streak characteristics (e.g. \cite{Li_2016_CVPR}\cite{Kim_2015_TIP}) to separate background and rain streak layers. Unfortunately, in the real world,  rain appearance does not  depend only on individual rain streaks, but also on the accumulation of multiple rain streaks in the space from the camera to the background scene, as shown in Fig.~\ref{fig:RainExample}.

In an image, the projected appearance of the rain streaks will have different sizes and densities. The further away the streaks, the smaller the size and the denser the imaged rain streaks. If we assume rain streaks at the same depth as one layer, we divide the captured rainy scene into a clean background and multiple layers of rain streaks, each of which has approximately the same size and density. Based on this, we can generalize Eq.~(\ref{eq:LinearSuperposition}) to model a rainy scene:
\begin{equation}
    \mathbf{O} = \mathbf{B} +  \sum_{i}^{n}\mathbf{R_i},
    \label{eq:LinearSuperpositionMultiple}
\end{equation}
where $n$ is the number of rain-streak layers along the line of sight to the background objects. \(\mathbf{R_i}\) represents the pixel intensity of rain streaks at layer $i$.

According to \cite{Kaushal2017}, rain droplets can cause light scattering and attenuation. Thus, the  resultant visibility under moderate and heavy rain conditions are similar to those under haze and fog. The light scattering process contributes to the atmospheric veiling effect in a typical heavy rainy scene (see blue window in Fig.~\ref{fig:RainExample}). In this case, the purely additive rain model introduced in Eq.~(\ref{eq:LinearSuperposition}) does not fully capture the appearance of rain accumulation, and as a result, the existing rain removal methods based on Eq.~(\ref{eq:LinearSuperposition}) cannot handle it.  To address this, we further generalize the rain model:
\begin{equation}
    \mathbf{O} = \boldsymbol{\alpha} \odot (\mathbf{B} + \sum_{i}^{n}\mathbf{R_i} ) + (\boldsymbol{1} - \boldsymbol{\alpha}) \odot \mathbf{A},
    \label{eq:RainStreakAccumulationEquation}
\end{equation}
where $\odot$ indicates element-wise multiplication,  \(\boldsymbol{\alpha}\) is a 2D map representing the transmittance introduced by the scattering process of rain droplets,  and $\mathbf{A}$ is a 2D map representing the atmospheric light of the scene. This model is similar to that proposed in  \cite{Yang_2017_CVPR}, although we remove the  binary mask in our model.

\section{Deraining Method}
\begin{figure*} [h!]
	\includegraphics[width=1.0\linewidth]{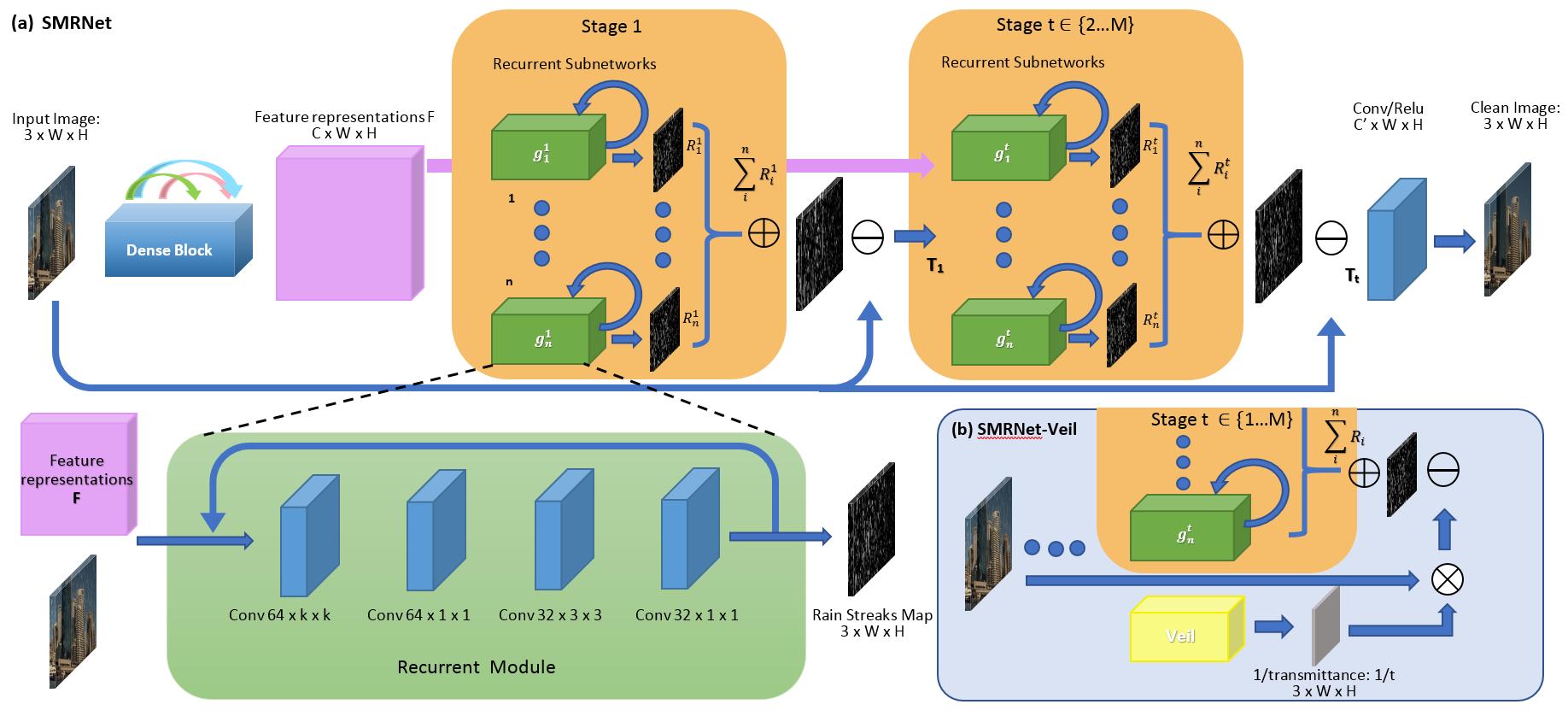}
	\caption{ The network architectures for SMRNet (a) and SMRNet-Veil (b).  }
	\label{fig:NetArch}
\end{figure*}
Our scale-aware multi-stage recurrent deraining network is illustrated in Fig.~\ref{fig:NetArch}.
Unlike the existing CNN-based deraining methods, we create multiple recurrent sub-networks to handle rain streaks at different sizes and densities.  The veiling effect of rain accumulation, $(\boldsymbol{1}-\boldsymbol{\alpha})\mathbf{A}$, can be deemed as another layer and thus can be processed using another recurrent sub-network in parallel to those that handle rain streaks. However, since the underlying physical process generating the veiling effect is different from that of rain streaks (Fig.~\ref{fig:RainExample} Green), the network architecture treats this effect differently. For ease of discussion, we will first discuss our network that deals only with rain streaks, called '\textbf{SMRNet}' (Scale-aware Multi-stage Recurrent Network) and shown in Fig.~\ref{fig:NetArch}(a). Then, we will discuss the integration of this network with the sub-network that deals with the veiling effect, which we call '\textbf{SMRNet-veil}', shown in Fig.~\ref{fig:NetArch}(b).

\subsection{SMRNet}
Focusing on multiple layers and overlapping rain streaks, based on Eq.~(\ref{eq:LinearSuperpositionMultiple}), our goal is to estimate the rain-free background \(\mathbf{B}\) and each rain streak intensity map \(\mathbf{R_1}, \mathbf{R_2}, ..., \mathbf{R_n} \) given the input rainy image \(\mathbf{O}\).  Generally, we want to minimize $\parallel \mathbf{O} - \mathbf{B} - \sum_{i}^{n}\mathbf{R_i} \parallel_F^2$,  where $\parallel \cdot \parallel_F$ is the Frobenius norm.  This is a totally ill-posed problem, even when we regard there is only one layer of rain streaks. Early rain streak removal methods like \cite{Li_2016_CVPR}\cite{chen2013generalized} use hand-crafted features or data-driven features for the priors of $\mathbf{B}$ and $\mathbf{R_i}$. However, in our method, the priors are learned by the network from the training data. In order to obtain the priors of $\mathbf{B}$ and $\mathbf{R_i}$ during the training phase, we add the estimation loss of $\mathbf{B}$ and $\mathbf{R_i}$ as in our objectives:
\begin{equation}
    \mathcal{L} = \mathcal{L_B} +  \sum_{i}^{n} \mathcal{L}_{\mathcal{R}_i},
    \label{eq:Lossfunction}
\end{equation}
where \(\mathcal{L_B}\) represents reconstruction loss for the background scene, and \(\mathcal{L}_{\mathcal{R}_i}\) represents the loss for estimating the $i^{th}$ rain streak layer.

As illustrated in Fig.~\ref{fig:NetArch}, our network first adopts DenseNet \cite{Huang_2017_CVPR} to extract rain image features \(\mathbf{F}\). However, we remove the 'transition layers' (1 conv layer followed by 1 pooling layer) between each 'Dense block' from the DenseNet so that our network does not downsample the image. The features are then fed into a series of parallel recurrent convolutional sub-networks to learn rain intensity map \(\mathbf{R_i}\) at different scales and densities. Since each recurrent sub-network focuses only on one type of rain streak feature \(\mathbf{R_i}\), resources can be dedicated toward these selected rain streaks without having to accommodate the competing demands of the different rain streak types in the network with their different desiderata for representation, yielding an enhanced learning of foreground items.
The recurrent module iterates four times in order to refine the estimation using the previous prediction combined with the feature representations. Then, the input image will subtract all the estimated rain intensity maps \(\mathbf{R_i^{1}}\)'s to obtain a temporary image \(\mathbf{T_1}\) , which is a preliminary de-rained result. This preliminary result is often marred by dark streak artifacts and contains residual rain streaks. The dark streaks arise possibly due to ``double removal'' by two parallel sub-networks. The reasons for the residual rain streaks are manifold.

Firstly, even in a synthetic rain image, an image region may contain multiple rain streaks stacked right on top of one another. It is difficult to remove all the rain streaks in one go; instead, we are much more likely to see the removal of only the nearest and thickest rain streaks, upon which the further and finer streaks are revealed. Secondly, in the real world rain sequences, there might be effects not modelled in the training. For instance, the rainfall is simply heavier, or there are local variation of density and direction not related to the depth factor (e.g. the variation around the table in Fig.~\ref{fig:Cover}). These effects would pose difficulties for a single-stage network solution, even with its parallel sub-networks. In view of the preceding issues, we send the preliminary derained result to the next stage together with the feature representation \(\mathbf{F}\) and the various estimated rain layers for further refinement. A predictor $\mathbf{R_i^{j}}$ in next stage can better remove the residual rain streaks because firstly the dominant signals have been removed. This argument can be understood in various senses: (1) in the case of coincident rain streaks mentioned before, the removal of the nearest (and thus brightest) rain streaks reveal the underlying fainter rain streaks; (2) local rain streaks that are inconsistent in density or direction with the global pattern are better detected and processed after the dominant global pattern is removed; (3) the rain streaks can also be dominated by the veiling effect to be discussed in the next subsection, and the latter’s removal in the first stage helps reveal rain streaks better.

The second reason for having these multiple stages is when we are faced with an unprecedented heavy rain not seen in the training. The first stage may only partially remove the rain streaks, having not seen such dense rain streak pattern before. However, the partially derained result at the end of the first stage amounts to an image of a lighter rainy scene, and we find that the successive stages can successfully remove the remaining rain streaks. The third reason is concerned with the removal of the dark streak artifacts. One can regard the concatenated predictions from the earlier stages as providing some form of explicit 'communication' between each recurrent sub-network, leading to reduced 'duplicate work'. At the end of the network, the learned rain streaks from all the stages are concatenated together to aid the final clean image recovery.

\begin{figure}
    \includegraphics[width=0.251\linewidth]{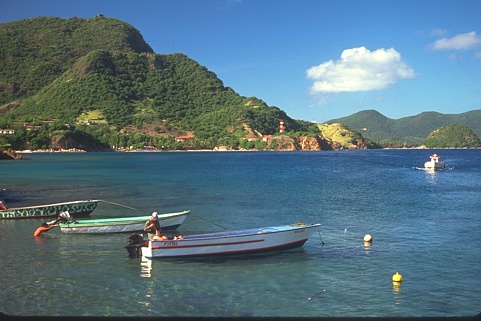}
    \includegraphics[width=0.251\linewidth]{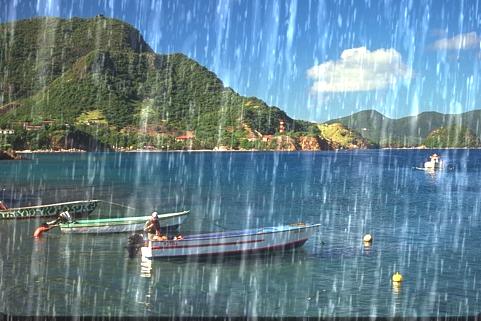}
    \includegraphics[width=0.223\linewidth]{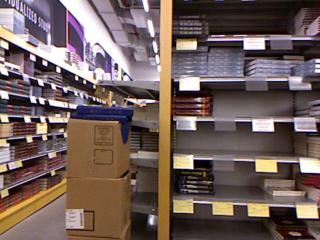}
    \includegraphics[width=0.223\linewidth]{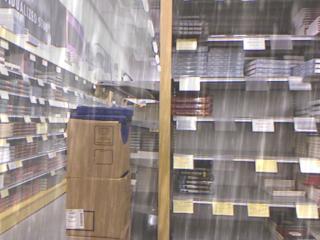}

    \includegraphics[width=0.32\linewidth]{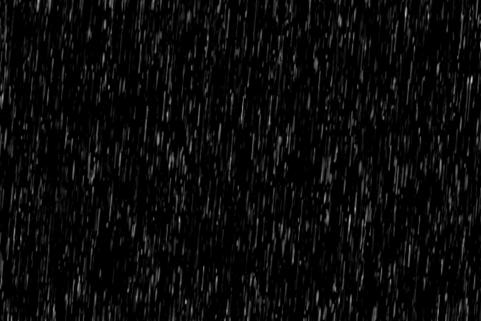}
    \includegraphics[width=0.32\linewidth]{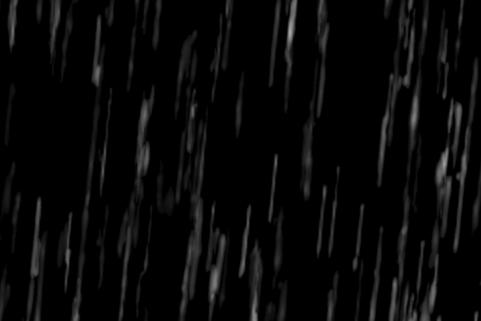}
    \includegraphics[width=0.32\linewidth]{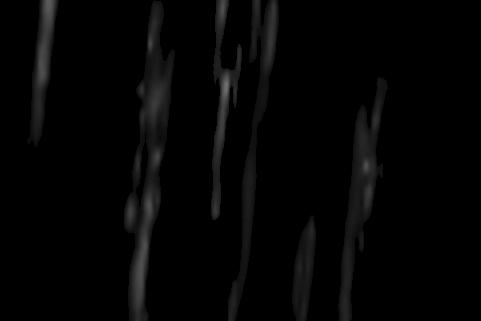}

    \includegraphics[width=0.32\linewidth]{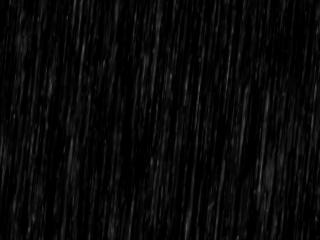}
    \includegraphics[width=0.32\linewidth]{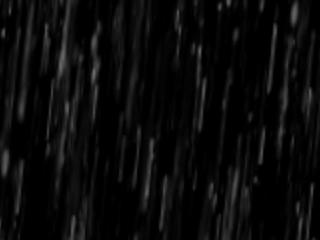}
    \includegraphics[width=0.32\linewidth]{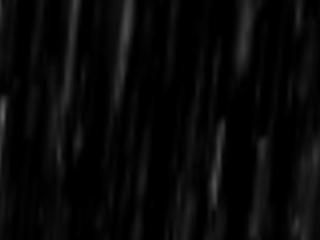}
    \caption{Examples of training data for SMR-Net (Top row left) and SMRNet-Veil (Top row right). \textbf{Middle} : Rain streak maps of the blue lake example (top left). \textbf{Bottom}: Rain streak maps of the indoor shelf (top right) example. From left to right are the corresponding maps for small-sized, mid-sized, and large-sized rain streaks respectively. (The rain streak map intensity is enhanced for visualization purpose) }
    \label{fig:Trainingdata}
\end{figure}

\setlength{\tabcolsep}{2pt}
\begin{table*}[]
\begin{center}
\caption{Quantitative results of different methods on synthetic rain datasets. }
\begin{tabular}{| c || c c c c| c c c c| c c c c| c c c c| }
\hline
Method    & \multicolumn{4}{c|}{Rain12 \cite{Li_2016_CVPR}\cite{Yang_2017_CVPR}}          & \multicolumn{4}{c|}{Rain12S}           & \multicolumn{4}{c|}{Rain100-COCO}       &\multicolumn{4}{c|}{Rain12-Veil}     \\
\hline
Metric                         & PSNR   & SSIM  & VIF    & FSIM  & PSNR   & SSIM  & VIF   & FSIM  & PSNR   & SSIM  & VIF   & FSIM   & PSNR  & SSIM  & VIF   & FSIM\\
\hline
ID  \cite{Kang12Rain}          & 27.21  & 0.800 & 0.266  & 0.765 & 25.07  & 0.773 & 0.229 & 0.757 & 20.91  & 0.667 & 0.24  & 0.785  & 18.34 & 0.630 & 0.179 & 0.704\\
\hline
DSC \cite{Luo_SparseCoding}    & 30.02  & 0.893 & 0.548  & 0.918 & 28.37  & 0.860 & 0.463 & 0.882 & 24.44  & 0.762 & 0.415 & 0.836  & 21.77 & 0.786 & 0.427 & 0.874 \\
\hline
LP  \cite{Li_2016_CVPR}        & 32.02  & 0.925 & 0.524  & 0.936 & 29.41  & 0.895 & 0.430 & 0.901 & 24.54  & 0.807 & 0.409 & 0.876  & 20.42 & 0.750 & 0.365 & 0.854 \\
\hline
Details Net \cite{Fu_2017_CVPR}& 33.43  & 0.949 & 0.608  & 0.955 & 31.01  & 0.931 & 0.507 & 0.934 & 26.48  & 0.838 & 0.438 & 0.899  & 22.24 & 0.831 & \textbf{0.458} & \textbf{0.921} \\
\hline
JORDER \cite{Yang_2017_CVPR}   & 35.86  & 0.956 & 0.627  & 0.963 & 29.69  & 0.913 & 0.472 & 0.922 & 25.79  & 0.823 & 0.416 & 0.894  & 20.34 & 0.788 & 0.417 & 0.898 \\
\hline
JORDER-R \cite{Yang_2017_CVPR} & 36.02  & 0.934\footnotemark[\value{footnote}] & 0.516\footnotemark[\value{footnote}]  & 0.937\footnotemark[\value{footnote}] & 28.25  & 0.889 & 0.396 & 0.899 & 25.16  & 0.801 & 0.352 & 0.878  & 20.20 & 0.768 & 0.357 & 0.880 \\
\hline
\hline
Ours     & \textbf{36.23}  & \textbf{0.965} & \textbf{0.636}  & \textbf{0.968} & \textbf{32.82}  & \textbf{0.949} & \textbf{0.540} & \textbf{0.952} & \textbf{29.66}  & \textbf{0.877} & \textbf{0.490} & \textbf{0.934}  & \textbf{25.29}    & \textbf{0.843}  & 0.417       & 0.884 \\
\hline
\hline
\end{tabular}
\label{table:SyntheticQuantitativeResult}
\end{center}
\end{table*}

\subsection{Veil Module}
Not only does the veiling effect, degrade visibility, its presence also hinders the complete removal of all the rain streaks effectively, as shown in Fig.~\ref{fig:Overlap}.  For these reasons, we develop an additional module to specifically handle it, with the module placed in parallel with the recurrent rain streak sub-networks. Taking the rain accumulation transmittance \(\boldsymbol{\alpha}\) from Eq.~(\ref{eq:RainStreakAccumulationEquation}) into consideration, the new data term takes on the form of $ \parallel \mathbf{O} - \boldsymbol{\alpha}(\mathbf{B} + \sum_{i}^{n}\mathbf{R_i}) - (\boldsymbol{1}-\boldsymbol{\alpha})\mathbf{A} \parallel_F^2$.

Rearranging Eq.~(\ref{eq:RainStreakAccumulationEquation}), $\mathbf{B}$ can be written as:
\begin{equation}
    \mathbf{B} = {\boldsymbol{1} \over \boldsymbol{\alpha}}(\mathbf{O} - \mathbf{A}) - \sum_{i}^{n}\mathbf{R_i} + \mathbf{A} .
    \label{eq:AdjustedRainModel}
\end{equation}
Based on Eq.~(\ref{eq:AdjustedRainModel}), the proposed network estimates \( 1 \over \boldsymbol{\alpha} \) instead of $\boldsymbol{\alpha}$ so that the clean background image $\mathbf{B}$ can be directly predicted by element-wise multiplication of $1 \over \boldsymbol{\alpha}$ and the input image $\mathbf{O}$. In this way, our network can be trained in an end-to-end manner. Hence, the corresponding loss function for SMRNet-veil model is:
\begin{equation}
    \mathcal{L} = \mathcal{L_B} + \mathcal{L}_{\boldsymbol{1} \over \boldsymbol{\alpha}} +  \sum_{i}^{n} \mathcal{L}_{\mathcal{R}_i},
    \label{eq:LossfunctionVeil}
\end{equation}
where \(\mathcal{L}_{{\boldsymbol{1}} \over {\boldsymbol{\alpha}}}\) indicates the loss for the \( \boldsymbol{1} \over \boldsymbol{\alpha} \) transmission map. In the training phase, we augment the training set with different values of $\mathbf{A}$ and assume these are known. In the test phase, we follow \cite{dehaze} and use the brightest pixel as $\mathbf{A}$.

Note that, in contrast to our method, previous methods \cite{dehaze,Yang_2017_CVPR} estimate the transmittance map and background scene separately, requiring a different process or network  to deal with the veiling effect, and thus cannot be trained end-to-end.

\footnotetext{indicates the results run by ourselves}
\section{Experiment}
\begin{figure*}
\centering
\setlength{\tabcolsep}{1.0pt}
\begin{tabular}{@{} ccccccc @{}}
 & Input & LP\cite{Li_2016_CVPR} & JORDER \cite{Yang_2017_CVPR} &  DetailsNet \cite{Fu_2017_CVPR} & Ours & Ground Truth  \\
    a &
    \begin{subfigure}{.16\textwidth}
	    \centering
	    \includegraphics[width=1.0\linewidth]{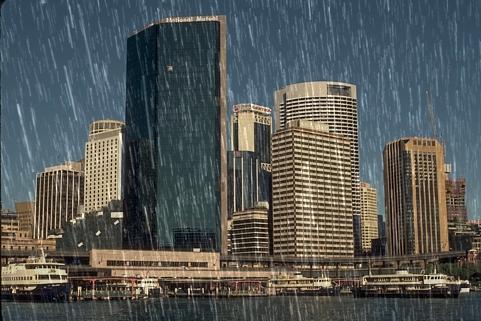}
	    \vspace{-0.35cm}
	\end{subfigure} &
    \begin{subfigure}{.16\textwidth}
	    \centering
	    \includegraphics[width=1.0\linewidth]{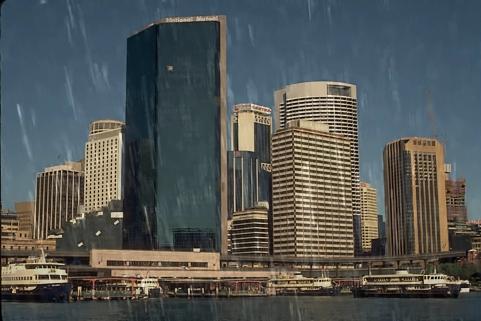}
	    \vspace{-0.35cm}
	\end{subfigure} &
    \begin{subfigure}{.16\textwidth}
	    \centering
	    \includegraphics[width=1.0\linewidth]{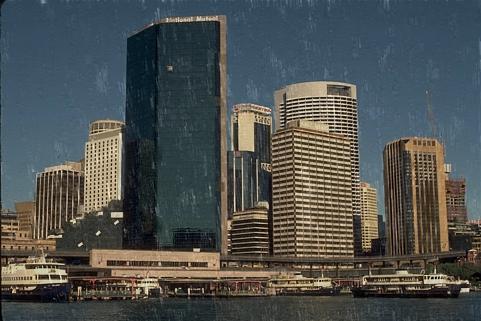}
	    \vspace{-0.35cm}
	\end{subfigure} &
	\begin{subfigure}{.16\textwidth}
	    \centering
	    \includegraphics[width=1.0\linewidth]{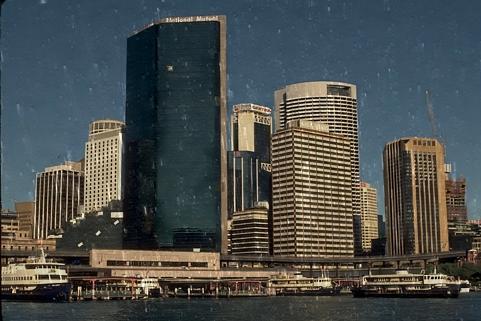}
	    \vspace{-0.35cm}
	\end{subfigure} &
	\begin{subfigure}{.16\textwidth}
	    \centering
	    \includegraphics[width=1.0\linewidth]{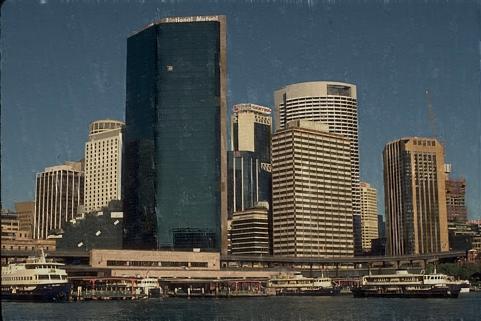}
	    \vspace{-0.35cm}
	\end{subfigure} &
    \begin{subfigure}{.16\textwidth}
	    \centering
	    \includegraphics[width=1.0\linewidth]{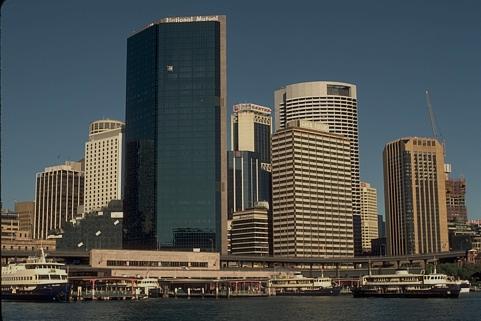}
	    \vspace{-0.35cm}
	\end{subfigure}
\\	
    b &
    \begin{subfigure}{.16\textwidth}
	    \centering
	    \includegraphics[width=1.0\linewidth]{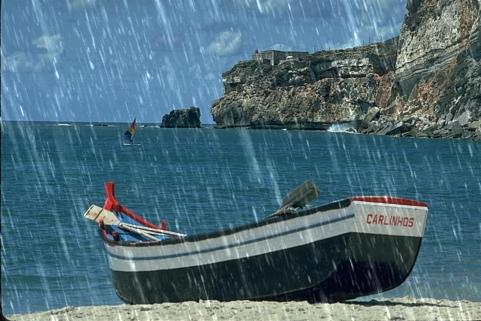}
	    \vspace{-0.35cm}
	\end{subfigure} &
    \begin{subfigure}{.16\textwidth}
	    \centering
	    \includegraphics[width=1.0\linewidth]{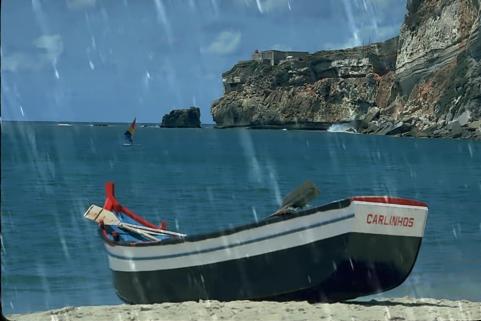}
	    \vspace{-0.35cm}
	\end{subfigure} &
    \begin{subfigure}{.16\textwidth}
	    \centering
	    \includegraphics[width=1.0\linewidth]{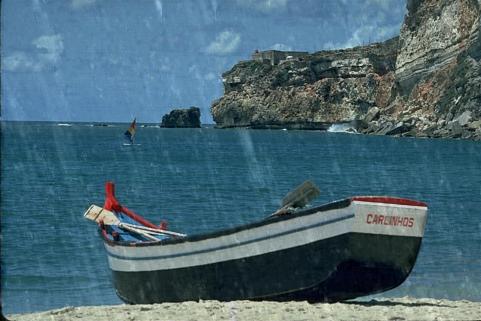}
	    \vspace{-0.35cm}
	\end{subfigure} &
	\begin{subfigure}{.16\textwidth}
	    \centering
	    \includegraphics[width=1.0\linewidth]{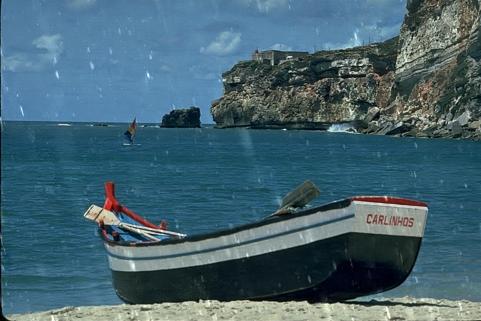}
	    \vspace{-0.35cm}
	\end{subfigure} &
	\begin{subfigure}{.16\textwidth}
	    \centering
	    \includegraphics[width=1.0\linewidth]{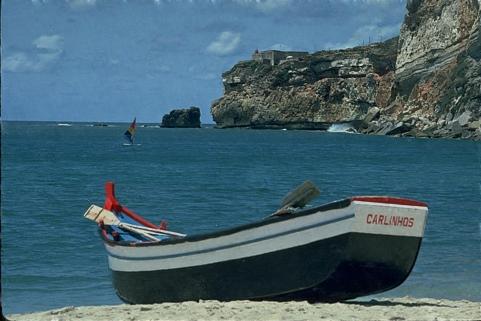}
	    \vspace{-0.35cm}
	\end{subfigure} &
    \begin{subfigure}{.16\textwidth}
	    \centering
	    \includegraphics[width=1.0\linewidth]{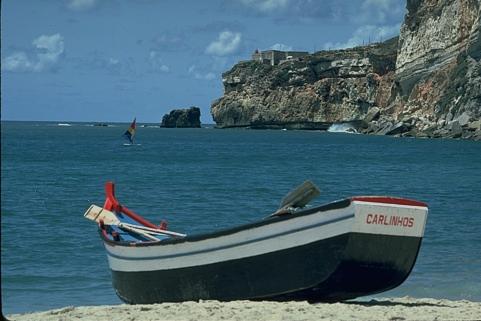}
	    \vspace{-0.35cm}

	\end{subfigure}
\\
    c &
    \begin{subfigure}{.16\textwidth}
	    \centering
	    \includegraphics[width=1.0\linewidth]{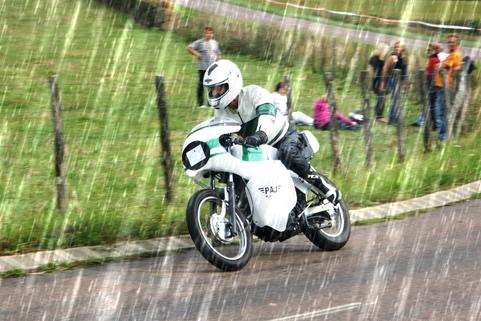}
	    \vspace{-0.35cm}
	\end{subfigure} &
    \begin{subfigure}{.16\textwidth}
	    \centering
	    \includegraphics[width=1.0\linewidth]{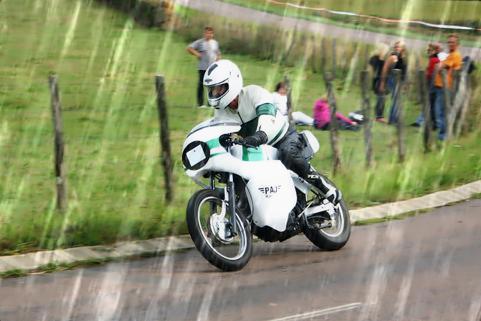}
	    \vspace{-0.35cm}
	\end{subfigure} &
    \begin{subfigure}{.16\textwidth}
	    \centering
	    \includegraphics[width=1.0\linewidth]{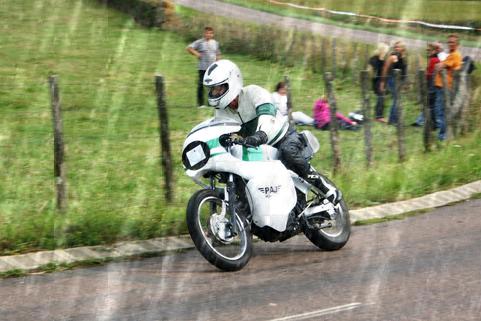}
	    \vspace{-0.35cm}
	\end{subfigure} &
    \begin{subfigure}{.16\textwidth}
	    \centering
	    \includegraphics[width=1.0\linewidth]{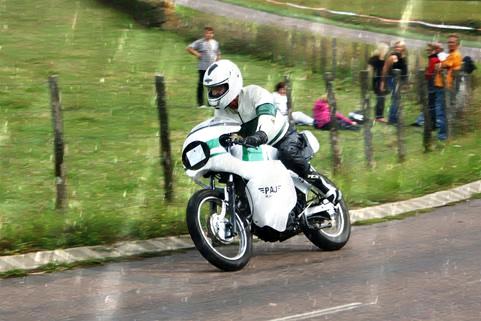}
	    \vspace{-0.35cm}
	\end{subfigure} &
	\begin{subfigure}{.16\textwidth}
	    \centering
	    \includegraphics[width=1.0\linewidth]{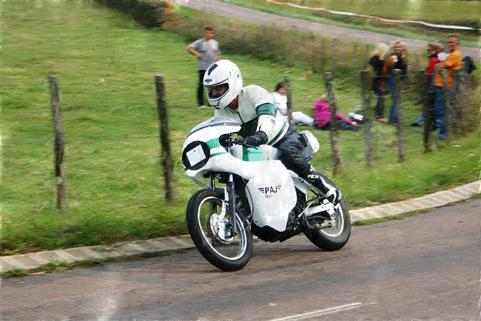}
	    \vspace{-0.35cm}
	\end{subfigure} &
	\begin{subfigure}{.16\textwidth}
	    \centering
	    \includegraphics[width=1.0\linewidth]{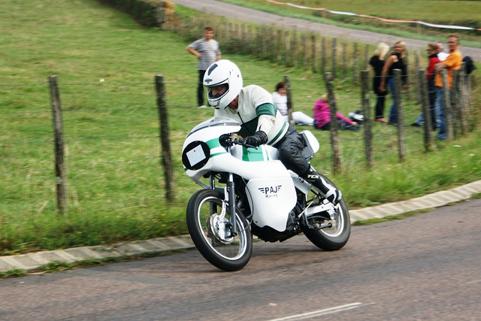}
	    \vspace{-0.35cm}
	\end{subfigure}
\\	
    d &
    \begin{subfigure}{.16\textwidth}
	    \centering
	    \includegraphics[width=1.0\linewidth]{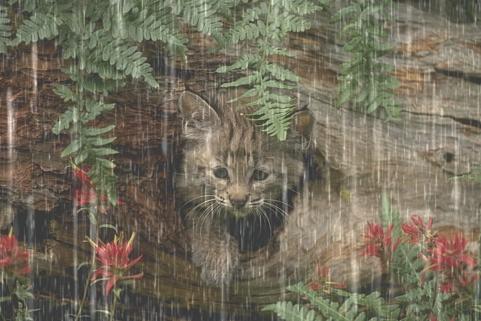}
	    \vspace{-0.35cm}
	\end{subfigure} &
    \begin{subfigure}{.16\textwidth}
	    \centering
	    \includegraphics[width=1.0\linewidth]{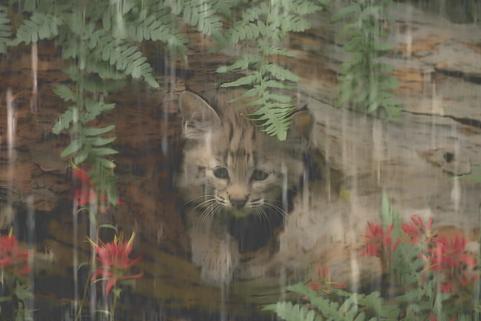}
	    \vspace{-0.35cm}
	\end{subfigure} &
    \begin{subfigure}{.16\textwidth}
	    \centering
	    \includegraphics[width=1.0\linewidth]{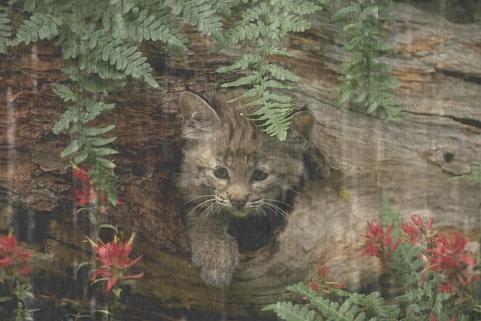}
	    \vspace{-0.35cm}
	\end{subfigure} &
    \begin{subfigure}{.16\textwidth}
	    \centering
	    \includegraphics[width=1.0\linewidth]{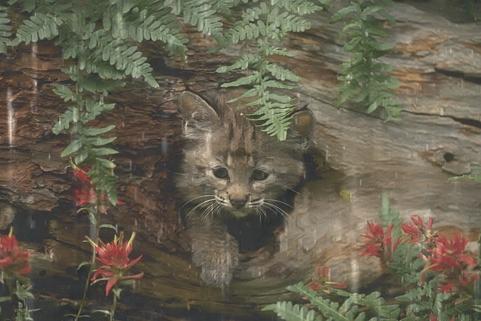}
	    \vspace{-0.35cm}
	\end{subfigure} &
	\begin{subfigure}{.16\textwidth}
	    \centering
	    \includegraphics[width=1.0\linewidth]{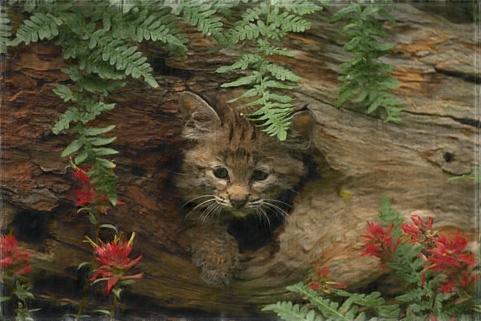}
	    \vspace{-0.35cm}
	\end{subfigure} &
	\begin{subfigure}{.16\textwidth}
	    \centering
	    \includegraphics[width=1.0\linewidth]{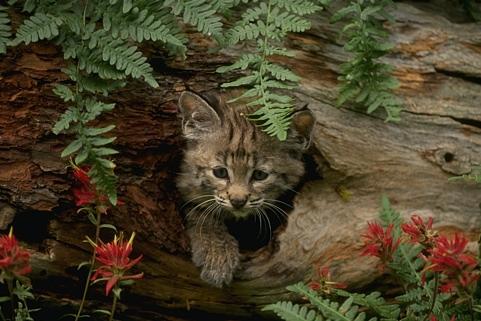}
	    \vspace{-0.35cm}
	\end{subfigure}
\\
\end{tabular}
\caption{Comparison of rain removal methods on synthetic rain images.  (\textbf{Best zoom in on screen}).}
\label{fig:SyntheticRainResult}
\end{figure*}

\subsection{Training Data}
In view of the difficulty in obtaining ground truths for real rain images, we choose to render synthesized rain streaks on clean natural images to fulfill the training need of our network. For testing, we use both synthetic and real rain data. Based on the model in \cite{Garg:2006}, we render synthesized rain streaks of multiple sizes and densities on the BSD300 \cite{BSD200} dataset; Eq.~(\ref{eq:LinearSuperpositionMultiple}) is used for the rendering for the training of SMRNet (see the left figure in Fig.~\ref{fig:Trainingdata}). These differently-sized rain streaks will be handled by different recurrent sub-networks. We divide these sizes into three ranges: 'small', 'middle', and 'large', respectively in the range of (0,60], (60,300], and (300, 600], where the size of a rain streak is measured by its occupied area (in pixel) on a rain image. In this experiment, we have synthesized 3300 rain images containing 11 different rain streak orientations.

For training SMRNet-Veil, we need depth information to render the veiling effect properly. Since the BSD300 dataset does not provide depth information, we use the NYU depth dataset \cite{Silberman:ECCV12} for this purpose. Specifically, we render the same rain streaks as before but with additional veiling effect generated according to Eq.~(\ref{eq:RainStreakAccumulationEquation}) (see the right figures in Fig.~\ref{fig:Trainingdata}). The transmittance $\alpha$ of a point $\mathbf{x}$ in the scene follows the free space light attenuation model \cite{Kaushal2017}:
\begin{equation}
    \alpha(\mathbf{x}) = \exp({- \beta d(\mathbf{x})}),
    \label{eq:LightAttenuation}
\end{equation}
where $d$ represents the depth of that point and the parameter $\beta$ is the attenuation factor. In generating different veiling effect, we set different values of $\beta$.

\subsection{Results of Synthetic Rain Data}
There are  four synthetic datasets evaluated in our experiments.  Table~\ref{table:SyntheticQuantitativeResult} shows the results of our method compared with other state-of-the-art rain streak removal methods.  To quantitatively evaluate these methods, four evaluation metrics are used: Peak Signal-to-Noise Ratio (PSNR) \cite{PSNR}, Structure Similarity Index (SSIM) \cite{SSIM}, Visual Information Fidelity (VIF) \cite{VIF_Sheikh_2004}, and Feature Similarity Index (FSIM) \cite{FSIM_zhang_2011}.

The Rain12 dataset \cite{Yang_2017_CVPR}\cite{Li_2016_CVPR} includes 12 synthesized rain images with only one type of rain streaks, which can be considered as light rain, as shown in Fig.~\ref{fig:SyntheticRainResult} (a). It is noteworthy that although our method does not focus on sparse light rain streak removal, its performance is still on par with the state-of-the-art performance. The Rain12S extends Rain12 dataset to include more adverse rain conditions, under which rain streaks have various sizes and densities as shown Fig.~\ref{fig:SyntheticRainResult} (b). From an inspection of the qualitative results, our method is able to remove all rain streaks of different sizes.  In order to compare the generalization of these methods, we also render differently-sized streaks on 100 images from the COCO dataset \cite{DBLP:journals/corr/LinMBHPRDZ14} using the same rain rendering method (shown in Fig.~\ref{fig:SyntheticRainResult} (c)). Our method outperforms other methods because it can handle the thick and thin rain streaks at the same time, thus restoring a clearer background. Finally, in order to evaluate the veiling effect removal, we render synthesized rain streaks and atmospheric veils following Eq.~(\ref{eq:RainStreakAccumulationEquation}) on 12 images from the BSD300 dataset (different from those used in the training data; see Fig.~\ref{fig:SyntheticRainResult}).

\begin{figure*}
\centering
\setlength{\tabcolsep}{1.0pt}
\begin{tabular}{@{} cccccc @{}}
& Input  & LP \cite{Li_2016_CVPR} & JORDER \cite{Yang_2017_CVPR} &  DetailsNet \cite{Fu_2017_CVPR} & Ours  \\
    a &
    \begin{subfigure}{.19\textwidth}
	    \centering
	    \includegraphics[width=1.0\linewidth]{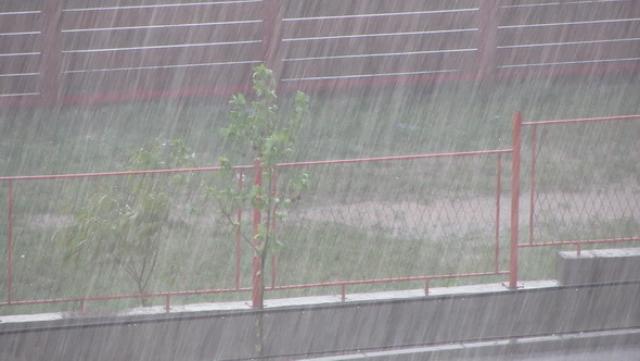}
	    \vspace{-0.35cm}
	\end{subfigure} &
    \begin{subfigure}{.19\textwidth}
	    \centering
	    \includegraphics[width=1.0\linewidth]{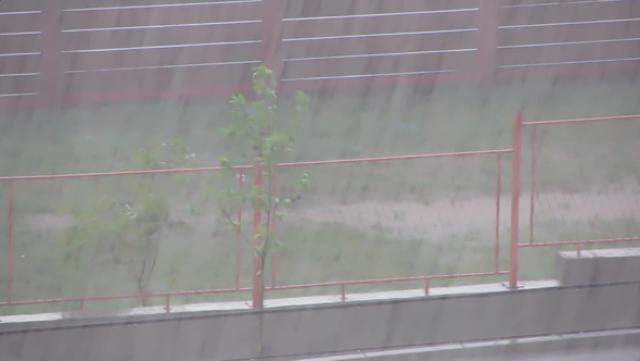}
	    \vspace{-0.35cm}
	\end{subfigure} &
    \begin{subfigure}{.19\textwidth}
	    \centering
	    \includegraphics[width=1.0\linewidth]{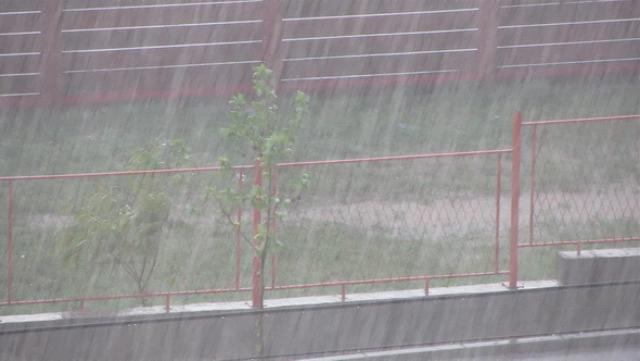}
	    \vspace{-0.35cm}
	\end{subfigure} &
	\begin{subfigure}{.19\textwidth}
	    \centering
	    \includegraphics[width=1.0\linewidth]{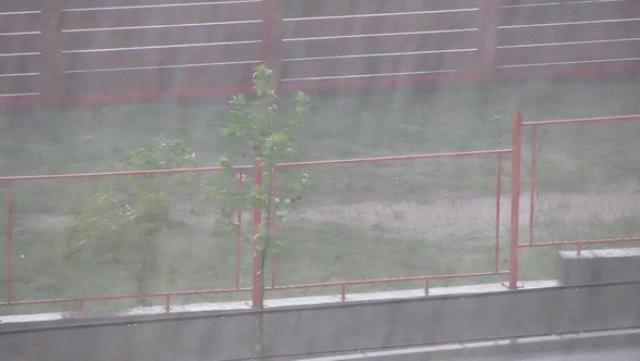}
	    \vspace{-0.35cm}
	\end{subfigure} &
	\begin{subfigure}{.19\textwidth}
	    \centering
	    \includegraphics[width=1.0\linewidth]{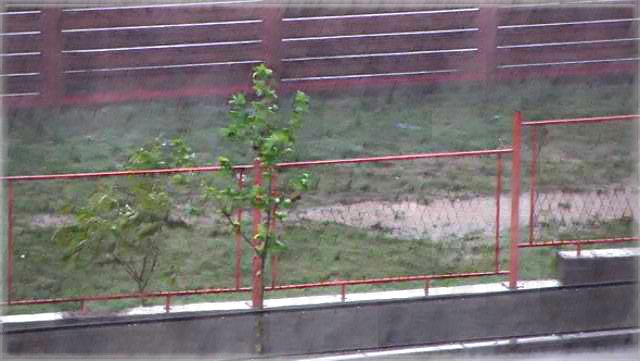}
	    \vspace{-0.35cm}
	\end{subfigure}
\\
    b &
    \begin{subfigure}{.19\textwidth}
	    \centering
	    \includegraphics[width=1.0\linewidth]{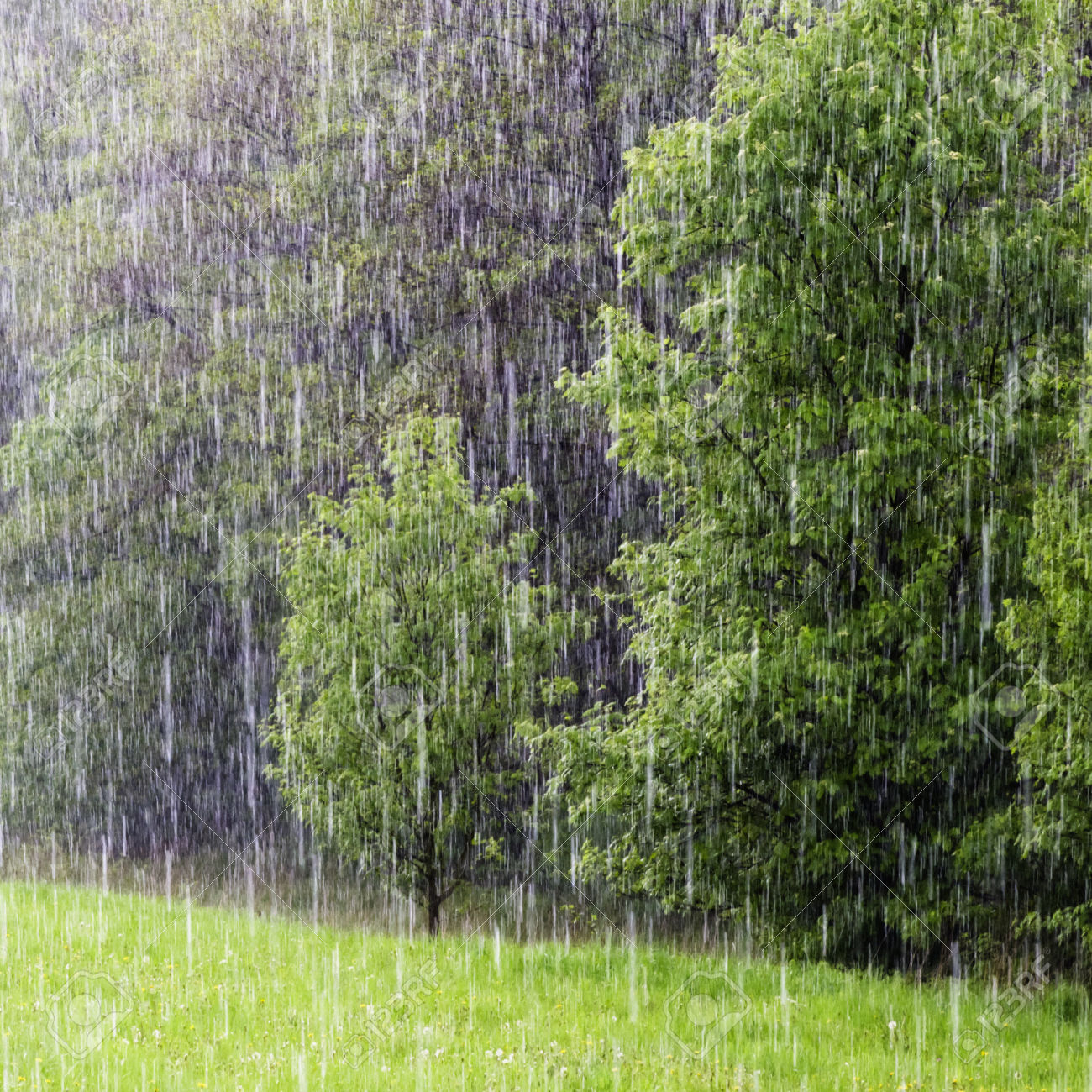}
	    \vspace{-0.35cm}
	\end{subfigure} &
    \begin{subfigure}{.19\textwidth}
	    \centering
	    \includegraphics[width=1.0\linewidth]{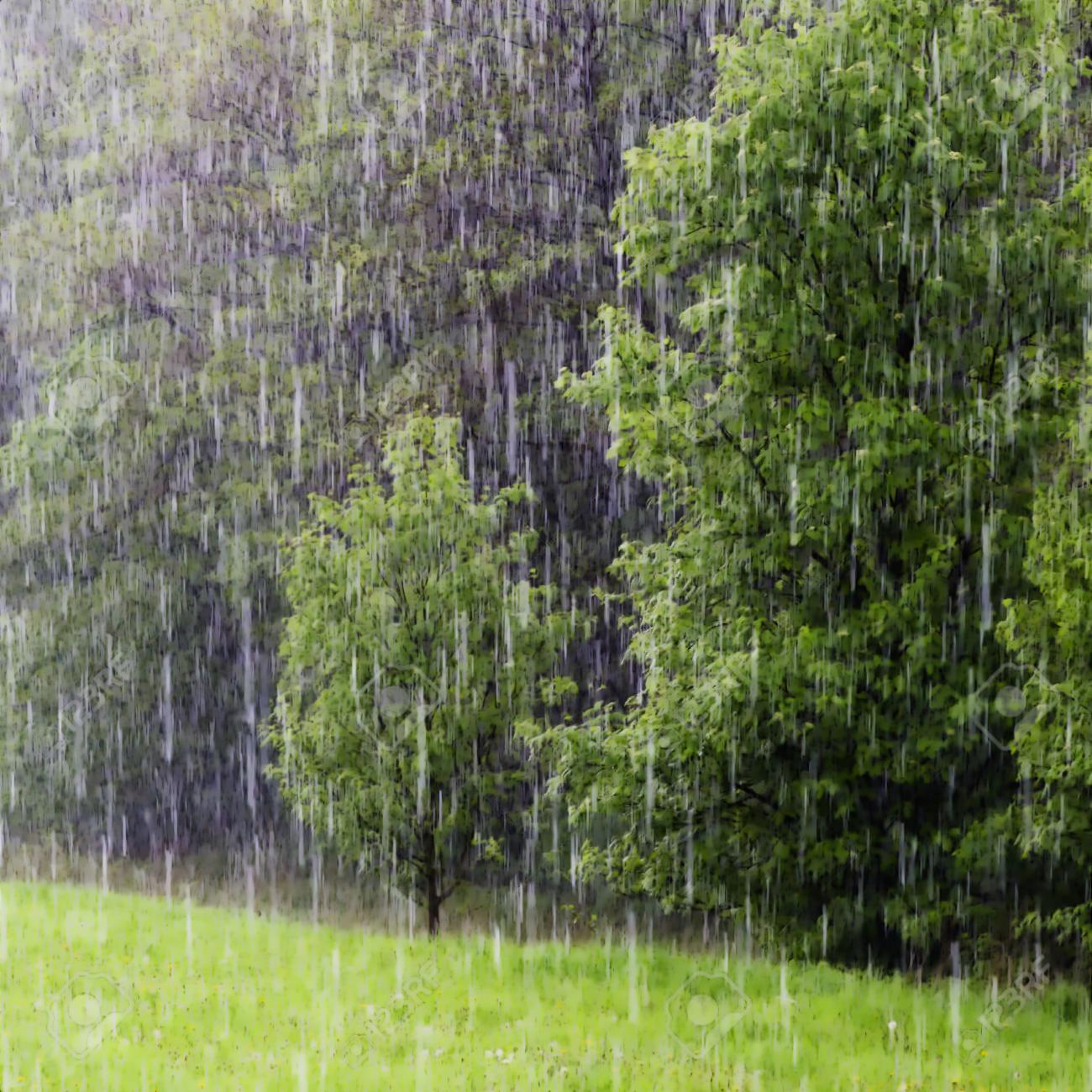}
	    \vspace{-0.35cm}
	\end{subfigure} &
    \begin{subfigure}{.19\textwidth}
	    \centering
	    \includegraphics[width=1.0\linewidth]{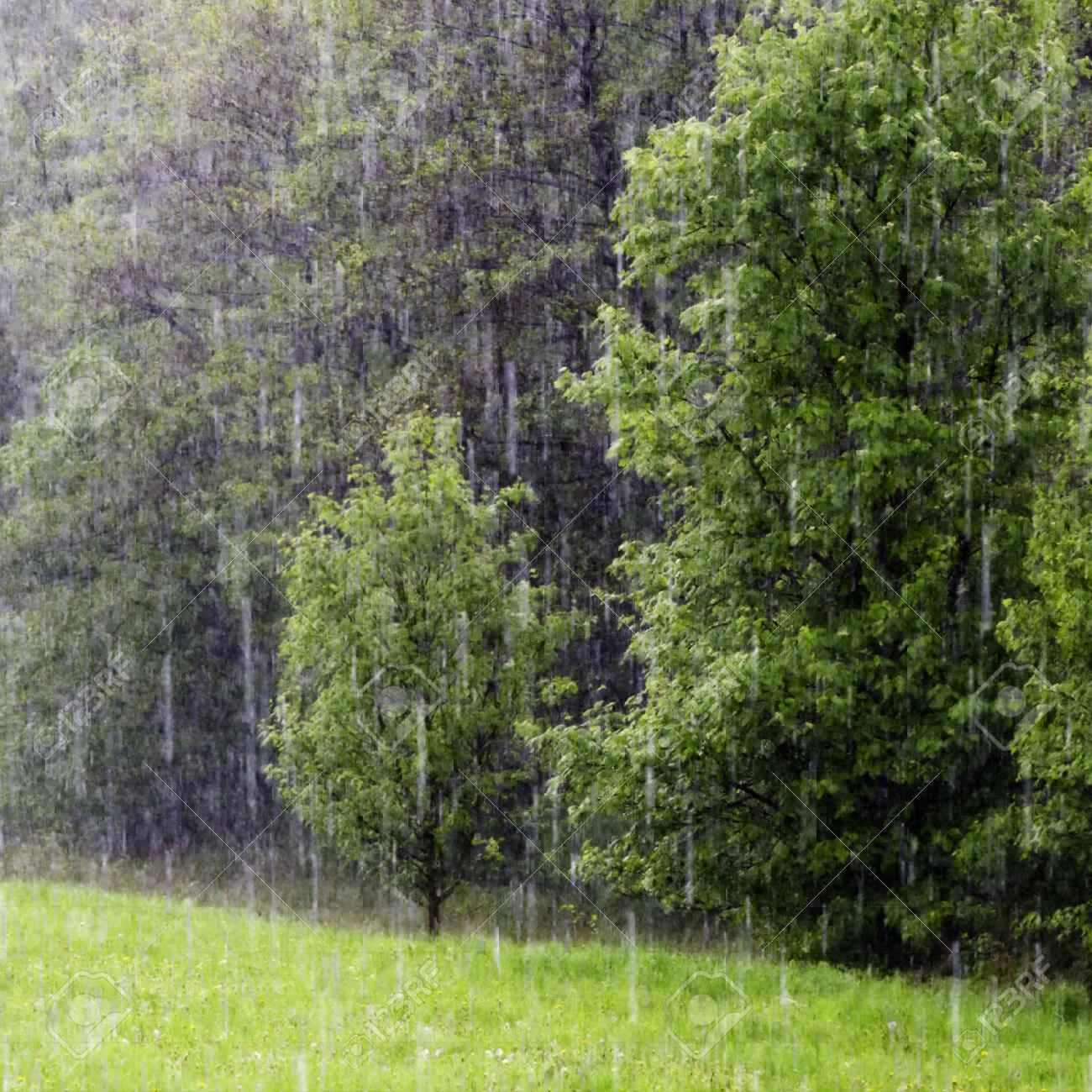}
	    \vspace{-0.35cm}
	\end{subfigure} &
	\begin{subfigure}{.19\textwidth}
	    \centering
	    \includegraphics[width=1.0\linewidth]{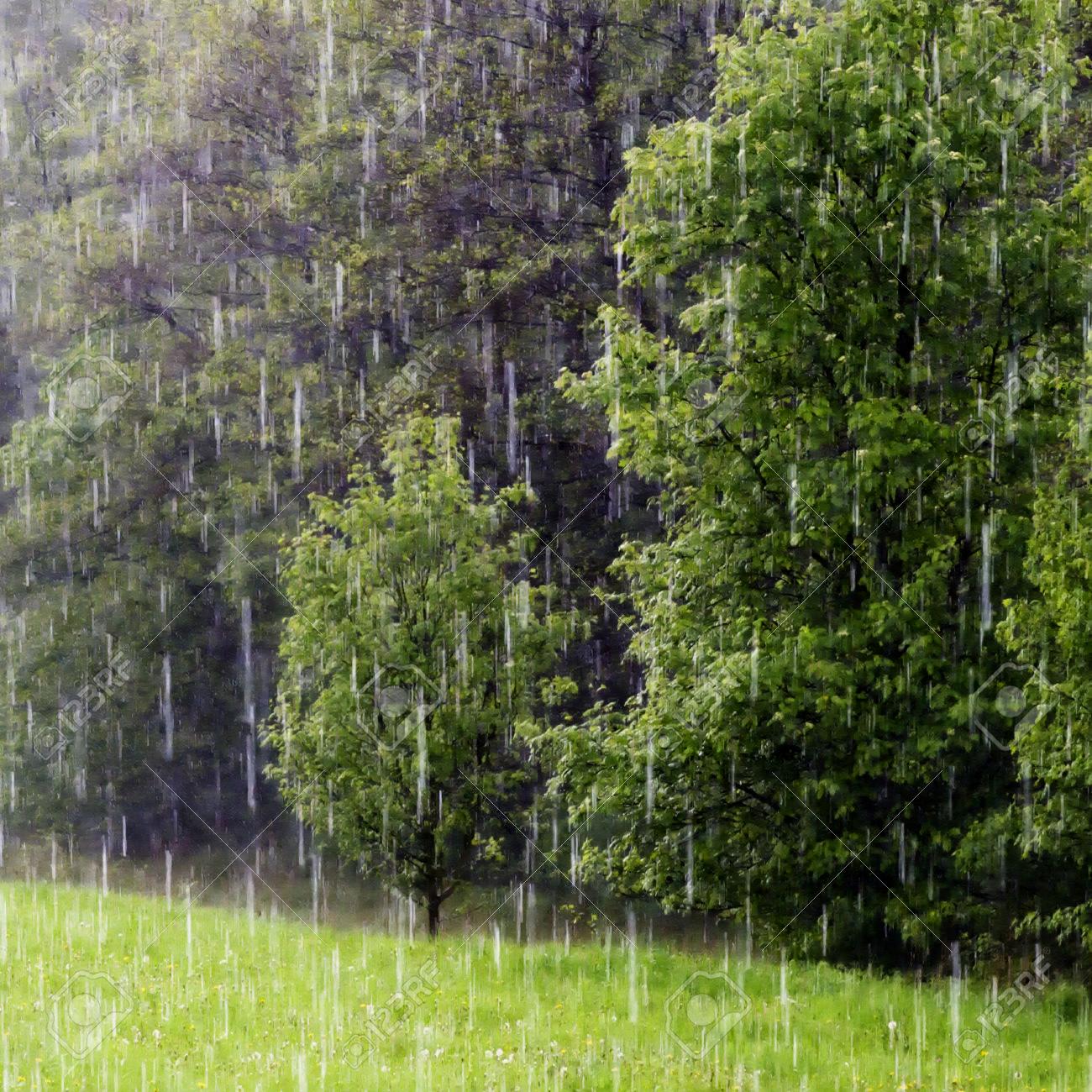}
	    \vspace{-0.35cm}
	\end{subfigure} &
	\begin{subfigure}{.19\textwidth}
	    \centering
	    \includegraphics[width=1.0\linewidth]{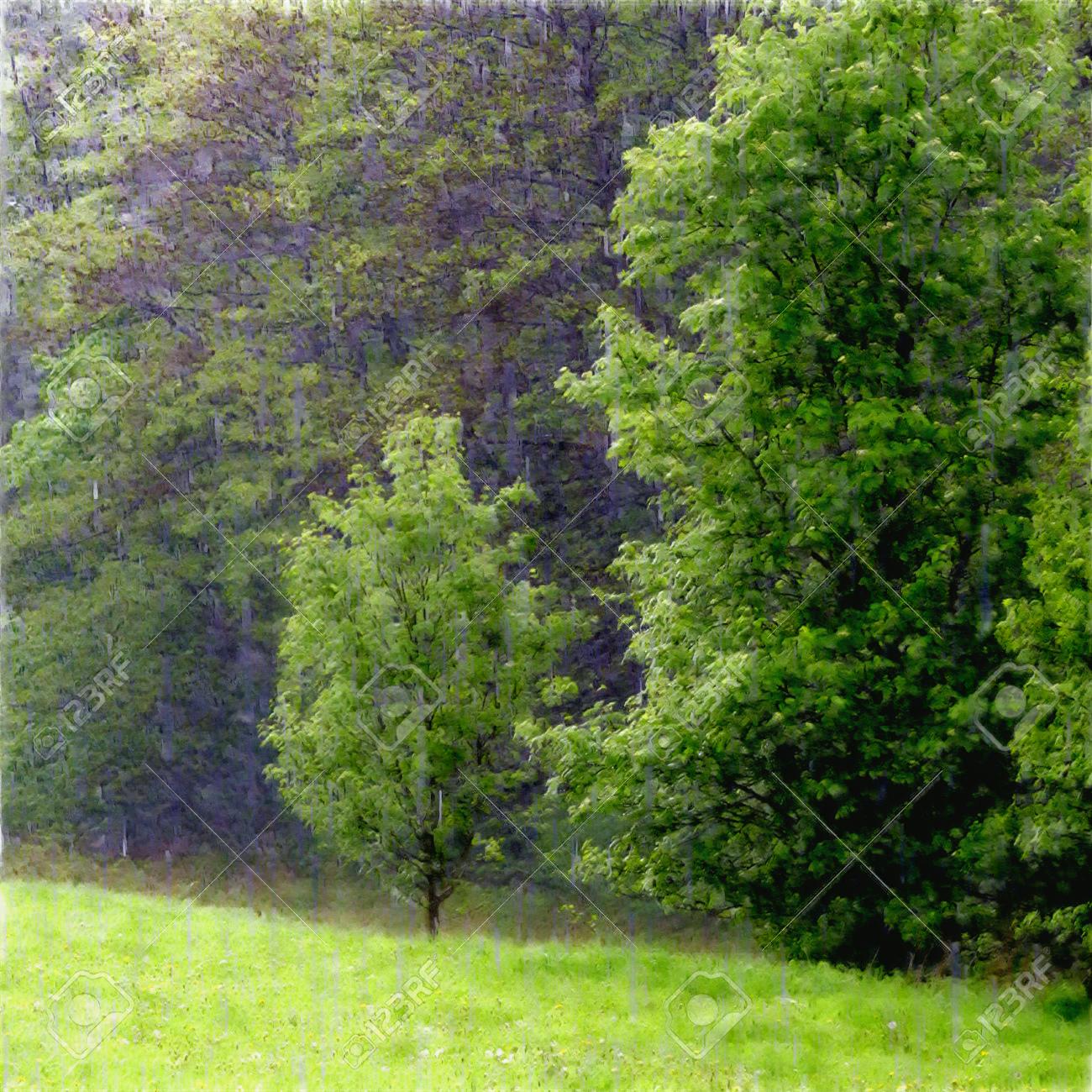}
	    \vspace{-0.35cm}
	\end{subfigure}
\\	
    c &
    \begin{subfigure}{.19\textwidth}
	    \centering
	    \includegraphics[width=1.0\linewidth]{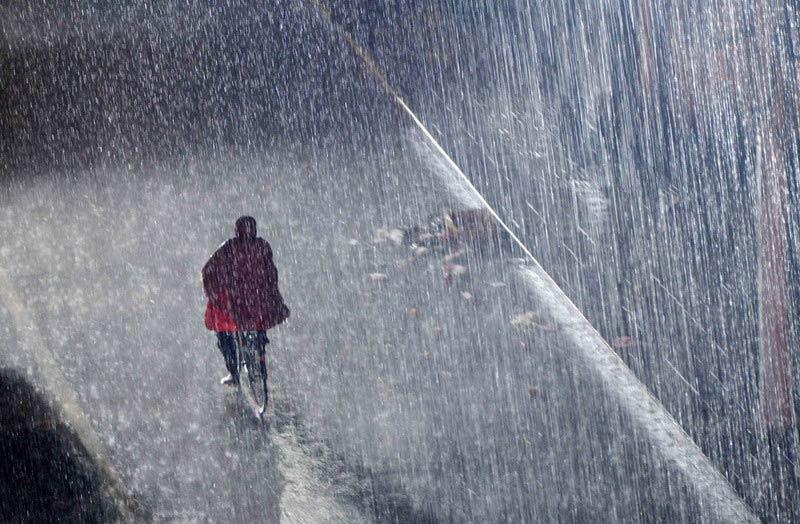}
	    \vspace{-0.35cm}
	\end{subfigure} &
    \begin{subfigure}{.19\textwidth}
	    \centering
	    \includegraphics[width=1.0\linewidth]{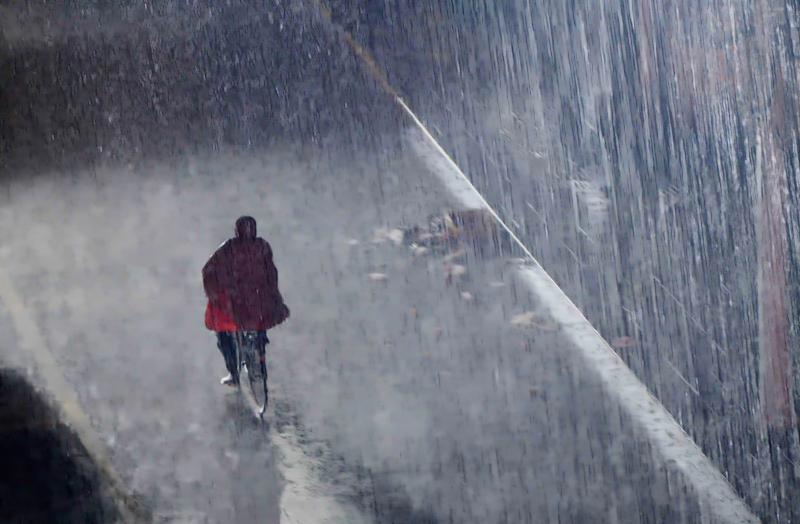}
	    \vspace{-0.35cm}
	\end{subfigure} &
    \begin{subfigure}{.19\textwidth}
	    \centering
	    \includegraphics[width=1.0\linewidth]{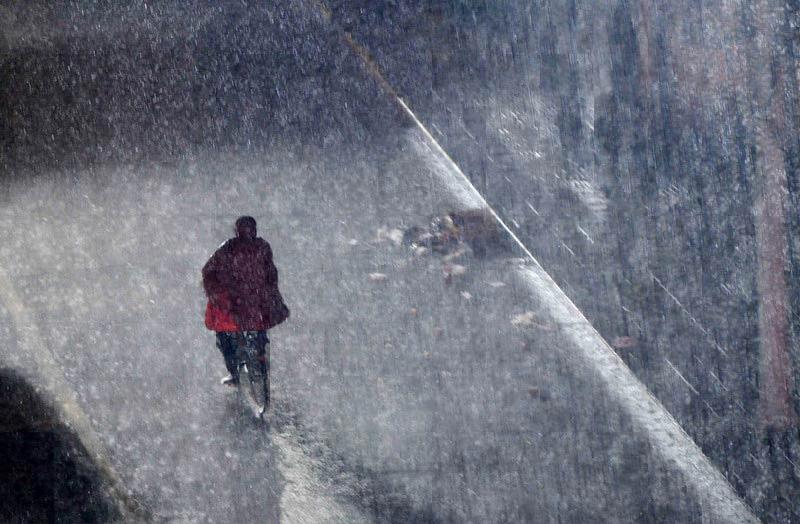}
	    \vspace{-0.35cm}
	\end{subfigure} &
	\begin{subfigure}{.19\textwidth}
	    \centering
	    \includegraphics[width=1.0\linewidth]{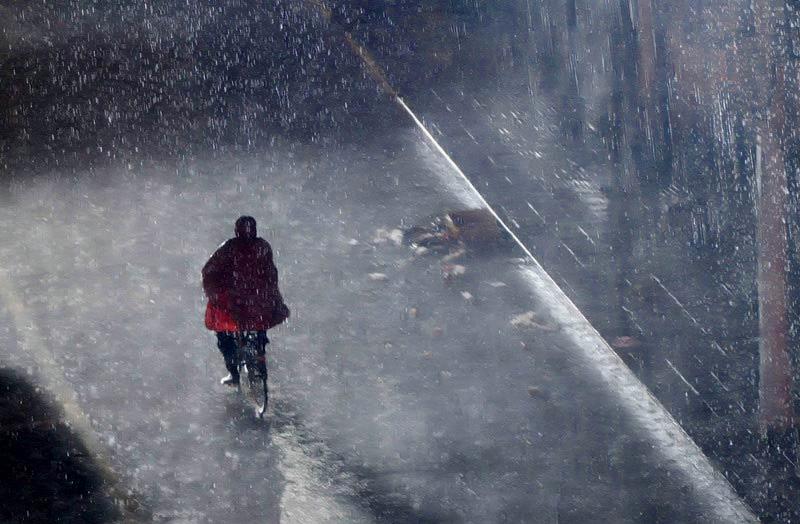}
	    \vspace{-0.35cm}
	\end{subfigure} &
	\begin{subfigure}{.19\textwidth}
	    \centering
	    \includegraphics[width=1.0\linewidth]{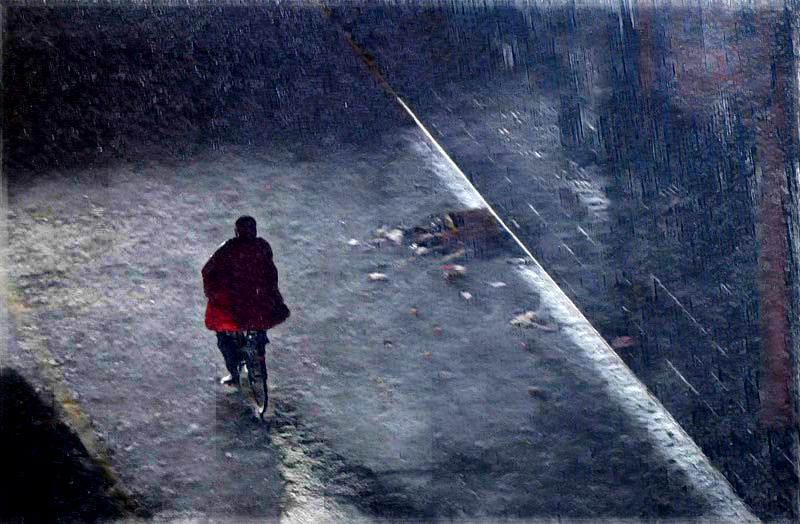}
	    \vspace{-0.35cm}
	\end{subfigure}
\\
    d &
    \begin{subfigure}{.19\textwidth}
	    \centering
	    \includegraphics[width=1.0\linewidth]{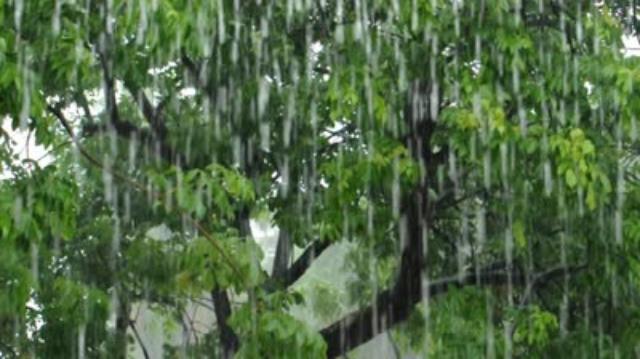}
	    \vspace{-0.35cm}
	\end{subfigure} &
    \begin{subfigure}{.19\textwidth}
	    \centering
	    \includegraphics[width=1.0\linewidth]{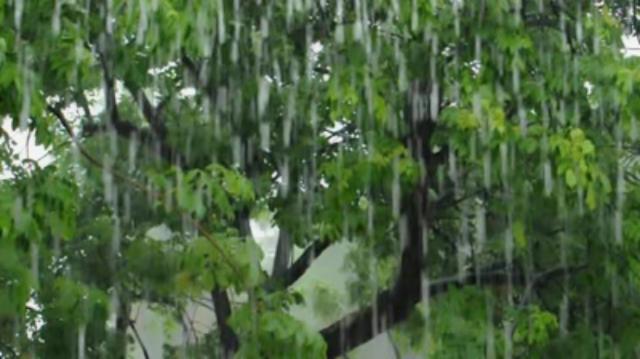}
	    \vspace{-0.35cm}
	\end{subfigure} &
    \begin{subfigure}{.19\textwidth}
	    \centering
	    \includegraphics[width=1.0\linewidth]{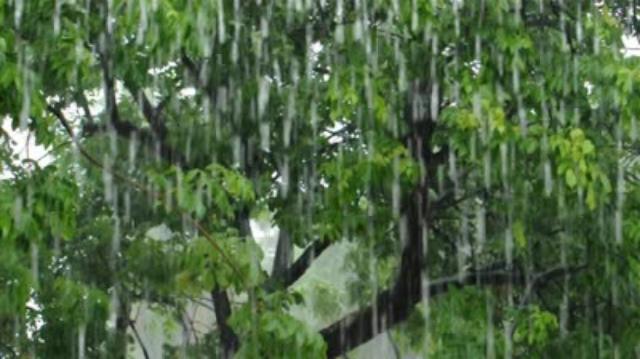}
	    \vspace{-0.35cm}
	\end{subfigure} &
	\begin{subfigure}{.19\textwidth}
	    \centering
	    \includegraphics[width=1.0\linewidth]{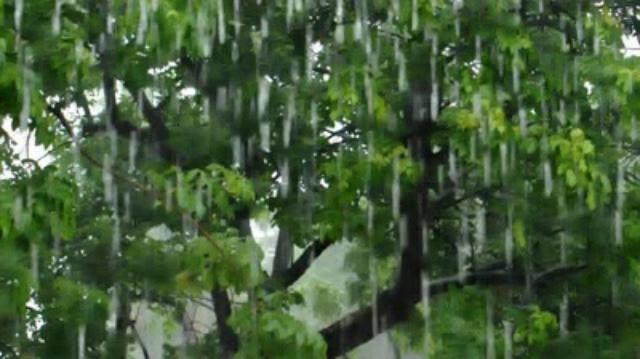}
	    \vspace{-0.35cm}
	\end{subfigure} &
	\begin{subfigure}{.19\textwidth}
	    \centering
	    \includegraphics[width=1.0\linewidth]{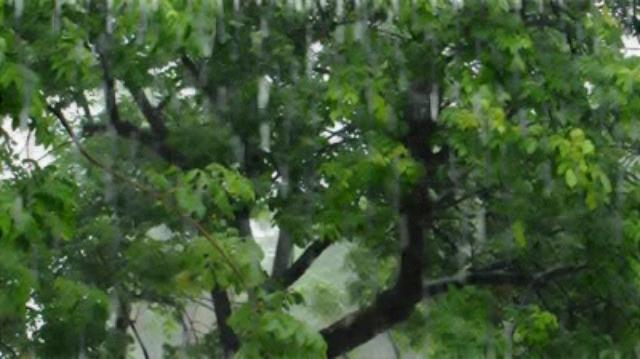}
	    \vspace{-0.35cm}
	\end{subfigure}
\\
    e &
    \begin{subfigure}{.19\textwidth}
	    \centering
	    \includegraphics[width=1.0\linewidth]{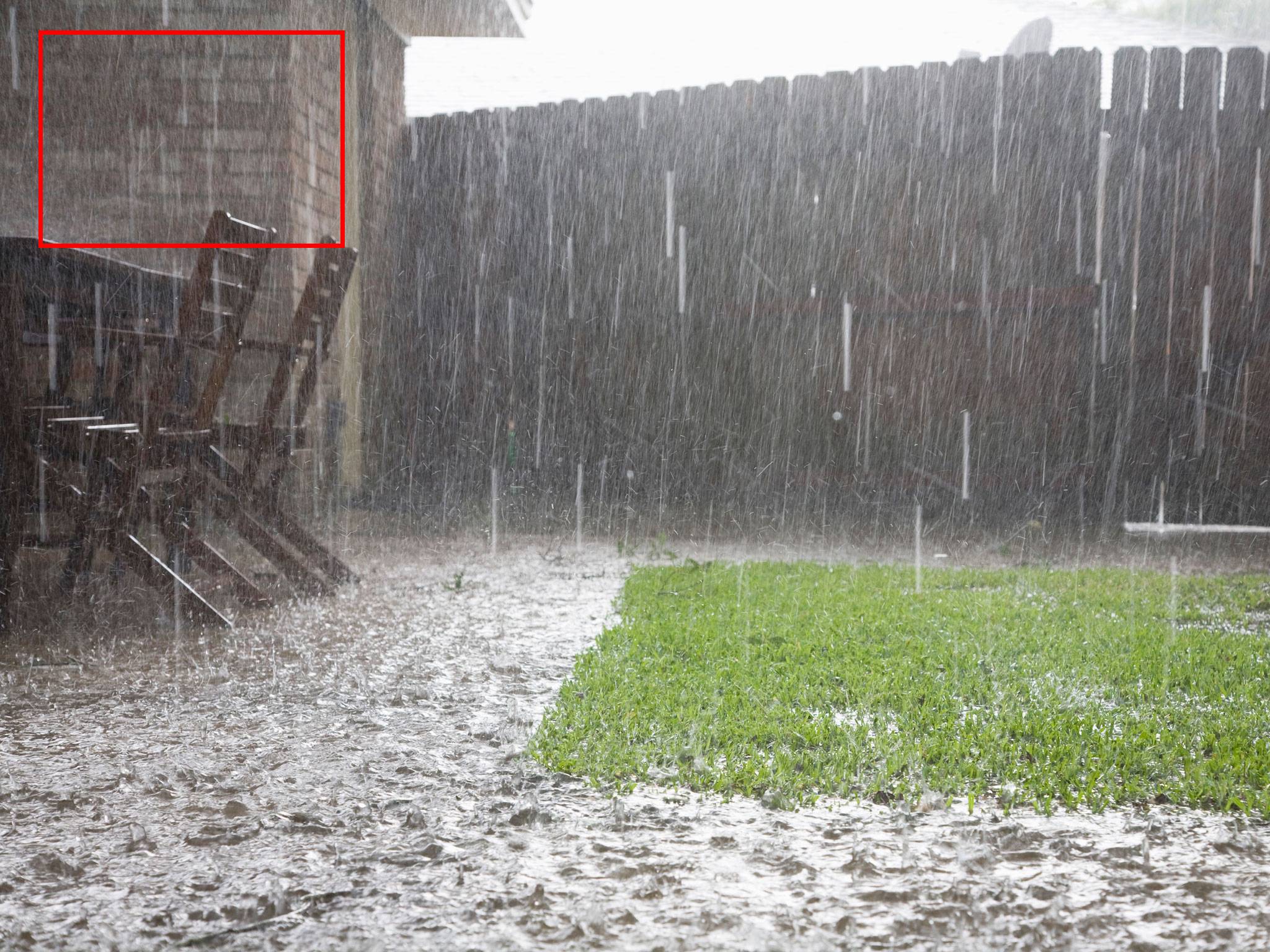}
	    \vspace{-0.35cm}
	\end{subfigure} &
    \begin{subfigure}{.19\textwidth}
	    \centering
	    \includegraphics[width=1.0\linewidth]{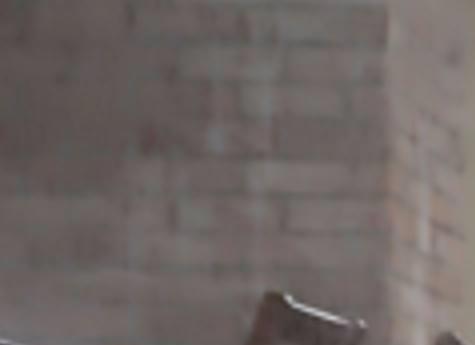}
	    \vspace{-0.35cm}
	\end{subfigure} &
    \begin{subfigure}{.19\textwidth}
	    \centering
	    \includegraphics[width=1.0\linewidth]{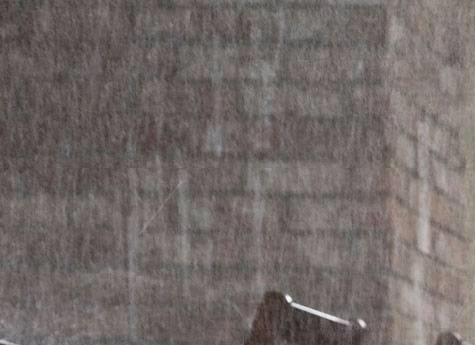}
	    \vspace{-0.35cm}
	\end{subfigure} &
	\begin{subfigure}{.19\textwidth}
	    \centering
	    \includegraphics[width=1.0\linewidth]{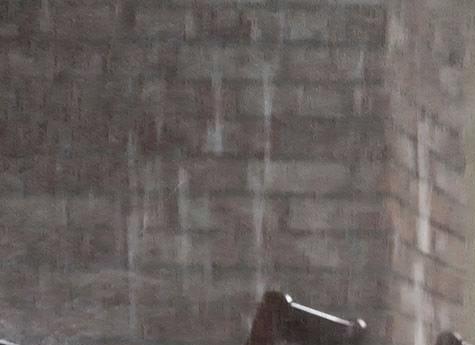}
	    \vspace{-0.35cm}
	\end{subfigure} &
	\begin{subfigure}{.19\textwidth}
	    \centering
	    \includegraphics[width=1.0\linewidth]{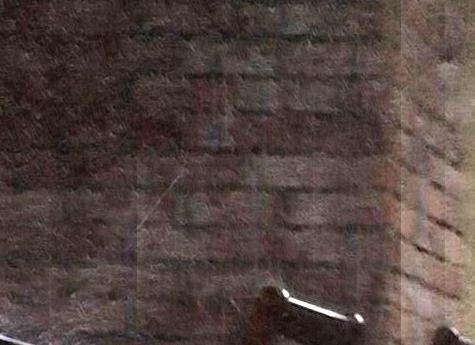}
	    \vspace{-0.35cm}
	\end{subfigure}
\\
\end{tabular}
\caption{Comparison of rain removal methods on real rain images. (e) shows the results of the red window of input image.}
\label{fig:RealRainResult}
\end{figure*}

\begin{figure}
    \includegraphics[width=0.98\linewidth]{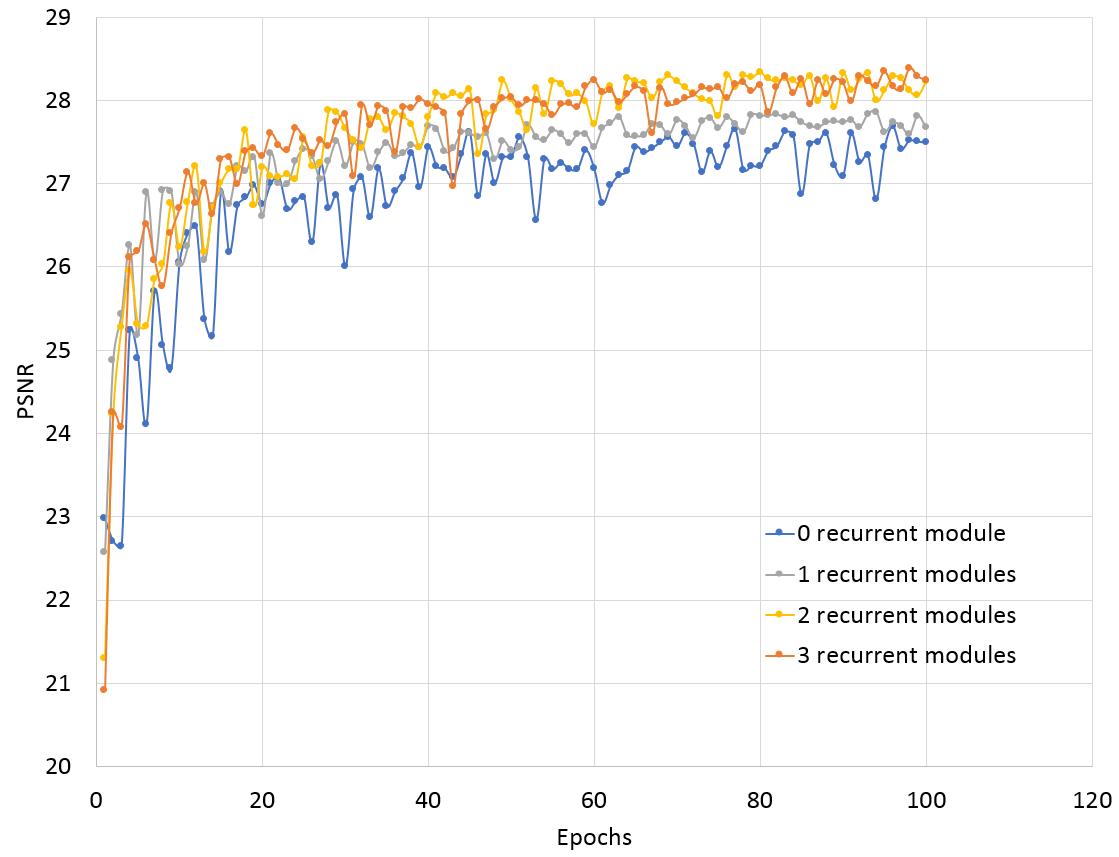}
    \caption{ The performance of the networks with different number of recurrent modules}
    \label{fig:CompareDifferentArch}
\end{figure}

\subsection{Results of Real Rain Data}
Fig.\ref{fig:RealRainResult} demonstrates the results of the proposed method and the recent state-of-the-art methods on real rain images. From the figure, one can see that our method is able to remove a range of different rain streaks, from dense and thin streaks to sparse and thick. However, JORDER \cite{Yang_2017_CVPR} cannot remove very thin nor very thick streaks (Fig.~\ref{fig:RealRainResult} (c)(d)). DetailsNet \cite{Fu_2017_CVPR} relies on the guided filter to separate high frequency signal from the real rain images. Low-frequency thick streaks in (d) cannot be fully extracted and therefore DetailsNet cannot remove them effectively. In the row (c), the rain streaks on the two sides of the image (zoom in to see clearly) are different and yet our method is able to remove all of them in one shot. For the cases with veiling effects (Fig.\ref{fig:RealRainResult} (a,b,c,e)), SMRNet-veil is able to recover the clean background without suffering from blurred object boundaries.

\subsection{Discussion}

\subsubsection{Evaluations on Parallel Modules}
In order to better understand the effectiveness of the network architecture, we compare a series of networks with different number of parallel recurrent modules, which are used to estimate the differently sized rain streaks. We denote the network that has no recurrent module but directly estimates the clean background as \textbf{0 recurrent module} network. The proposed network with 3 parallel recurrent modules is denoted as \textbf{3 recurrent modules} network. We evaluate these networks' performance at each epoch during training on the Rain100-COCO dataset. The performance results in PSNR metric are shown in Fig.~\ref{fig:CompareDifferentArch}. Due to the limited space, we include the ablation study of the effectiveness of recurrent module and the multiple stages in the supplementary material.


\section{Conclusion}
In this paper, we proposed a scale-aware multi-stage recurrent network to solve rain streak and rain streak accumulation removal problem. The proposed network estimates rain streaks of different sizes and densities individually in order to reduce intra-class competition. We generated multiple synthetic rain dataset to train our network and evaluate our network on both synthesized rain datasets and real rain dataset. The results demonstrate that our method attain the state-of-the-art performance.

{\small
\bibliographystyle{ieee}
\bibliography{egbib.bib}

\begin{thebibliography}{10}\itemsep=-1pt

\bibitem{Bossu2011}
J.~Bossu, N.~Hauti{\`e}re, and J.-P. Tarel.
\newblock Rain or snow detection in image sequences through use
  of a histogram of orientation of streaks.
\newblock {\em International Journal of Computer Vision}, 93(3):348--367, Jul
  2011.

\bibitem{dehaze}
B.~Cai, X.~Xu, K.~Jia, C.~Qing, and D.~Tao.
\newblock Dehazenet: An end-to-end system for single image haze removal.
\newblock {\em {IEEE} Trans. Image Processing}, 25(11):5187--5198, 2016.

\bibitem{chen2013generalized}
Y.-L. Chen and C.-T. Hsu.
\newblock A generalized low-rank appearance model for spatio-temporally
  correlated rain streaks.
\newblock In {\em Proceedings of the IEEE International Conference on Computer
  Vision}, pages 1968--1975, 2013.

\bibitem{DBLP:journals/corr/FuHDLP16}
X.~Fu, J.~Huang, X.~Ding, Y.~Liao, and J.~Paisley.
\newblock Clearing the skies: {A} deep network architecture for single-image
  rain removal.
\newblock {\em CoRR}, abs/1609.02087, 2016.

\bibitem{Fu_2017_CVPR}
X.~Fu, J.~Huang, D.~Zeng, Y.~Huang, X.~Ding, and J.~Paisley.
\newblock Removing rain from single images via a deep detail network.
\newblock In {\em The IEEE Conference on Computer Vision and Pattern
  Recognition (CVPR)}, July 2017.

\bibitem{Garg:2006}
K.~Garg and S.~K. Nayar.
\newblock Photorealistic rendering of rain streaks.
\newblock {\em ACM Trans. Graph.}, 25(3):996--1002, July 2006.

\bibitem{He_2015_CVPR}
K.~He, X.~Zhang, S.~Ren, and J.~Sun.
\newblock Deep residual learning for image recognition.
\newblock {\em CoRR}, abs/1512.03385, 2015.

\bibitem{Huang_2012_ICM}
D.~A. Huang, L.~W. Kang, M.~C. Yang, C.~W. Lin, and Y.~C.~F. Wang.
\newblock Context-aware single image rain removal.
\newblock In {\em 2012 IEEE International Conference on Multimedia and Expo},
  pages 164--169, July 2012.

\bibitem{Huang_2017_CVPR}
G.~Huang, Z.~Liu, L.~van~der Maaten, and K.~Q. Weinberger.
\newblock Densely connected convolutional networks.
\newblock In {\em The IEEE Conference on Computer Vision and Pattern
  Recognition (CVPR)}, July 2017.

\bibitem{PSNR}
Q.~Huynh-Thu and M.~Ghanbari.
\newblock Scope of validity of psnr in image/video quality assessment.
\newblock {\em Electronics Letters}, 44(13):800--801, June 2008.

\bibitem{Kang12Rain}
L.~W. Kang, C.~W. Lin, and Y.~H. Fu.
\newblock Automatic single-image-based rain streaks removal via image
  decomposition.
\newblock {\em IEEE Transactions on Image Processing}, 21(4):1742--1755, April
  2012.

\bibitem{Kaushal2017}
H.~Kaushal, V.~K. Jain, and S.~Kar.
\newblock {\em Free-Space Optical Channel Models}, pages 41--89.
\newblock Springer India, New Delhi, 2017.

\bibitem{Kim_2015_TIP}
J.~H. Kim, J.~Y. Sim, and C.~S. Kim.
\newblock Video deraining and desnowing using temporal correlation and low-rank
  matrix completion.
\newblock {\em IEEE Transactions on Image Processing}, 24(9):2658--2670, Sept
  2015.

\bibitem{Li_2016_CVPR}
Y.~Li, R.~T. Tan, X.~Guo, J.~Lu, and M.~S. Brown.
\newblock Rain streak removal using layer priors.
\newblock In {\em The IEEE Conference on Computer Vision and Pattern
  Recognition (CVPR)}, June 2016.

\bibitem{DBLP:journals/corr/LinMBHPRDZ14}
T.~Lin, M.~Maire, S.~J. Belongie, L.~D. Bourdev, R.~B. Girshick, J.~Hays,
  P.~Perona, D.~Ramanan, P.~Doll{\'{a}}r, and C.~L. Zitnick.
\newblock Microsoft {COCO:} common objects in context.
\newblock {\em CoRR}, abs/1405.0312, 2014.

\bibitem{Luo_SparseCoding}
Y.~Luo, Y.~Xu, and H.~Ji.
\newblock Removing rain from a single image via discriminative sparse coding.
\newblock In {\em 2015 IEEE International Conference on Computer Vision
  (ICCV)}, pages 3397--3405, Dec 2015.

\bibitem{BSD200}
D.~Martin, C.~Fowlkes, D.~Tal, and J.~Malik.
\newblock A database of human segmented natural images and its application to
  evaluating segmentation algorithms and measuring ecological statistics.
\newblock In {\em Proceedings Eighth IEEE International Conference on Computer
  Vision. ICCV 2001}, volume~2, pages 416--423 vol.2, 2001.

\bibitem{Silberman:ECCV12}
P.~K. Nathan~Silberman, Derek~Hoiem and R.~Fergus.
\newblock Indoor segmentation and support inference from rgbd images.
\newblock In {\em ECCV}, 2012.

\bibitem{VIF_Sheikh_2004}
H.~R. Sheikh and A.~C. Bovik.
\newblock Image information and visual quality.
\newblock {\em IEEE Transactions on Image Processing}, 15(2):430--444, Feb
  2006.

\bibitem{Tamburo2014}
R.~Tamburo, E.~Nurvitadhi, A.~Chugh, M.~Chen, A.~Rowe, T.~Kanade, and S.~G.
  Narasimhan.
\newblock {\em Programmable Automotive Headlights}, pages 750--765.
\newblock Springer International Publishing, Cham, 2014.

\bibitem{DBLP:conf/iccv/TianN09}
Y.~Tian and S.~G. Narasimhan.
\newblock Seeing through water: Image restoration using model-based tracking.
\newblock In {\em {IEEE} 12th International Conference on Computer Vision,
  {ICCV} 2009, Kyoto, Japan, September 27 - October 4, 2009}, pages 2303--2310,
  2009.

\bibitem{SSIM}
Z.~Wang, A.~C. Bovik, H.~R. Sheikh, and E.~P. Simoncelli.
\newblock Image quality assessment: from error visibility to structural
  similarity.
\newblock {\em IEEE Transactions on Image Processing}, 13(4):600--612, April
  2004.

\bibitem{Yang_2017_CVPR}
W.~Yang, R.~T. Tan, J.~Feng, J.~Liu, Z.~Guo, and S.~Yan.
\newblock Deep joint rain detection and removal from a single image.
\newblock In {\em The IEEE Conference on Computer Vision and Pattern
  Recognition (CVPR)}, July 2017.

\bibitem{FSIM_zhang_2011}
L.~Zhang, L.~Zhang, X.~Mou, and D.~Zhang.
\newblock Fsim: A feature similarity index for image quality assessment.
\newblock {\em IEEE Transactions on Image Processing}, 20(8):2378--2386, Aug
  2011.

\end{thebibliography}
}

\end{document}